\documentclass[runningheads]{llncs}
\usepackage{macro}
\newif\ifarxiv
\def\ifnotarxiv{\unless\ifarxiv}
\arxivtrue
\ifarxiv
\newcommand{\extref}[1]{#1}
\else
\newcommand{\extref}[1]{the extended version of the paper~\cite{wu2024marabou}}
\fi

\begin{document}
\title{Marabou 2.0: A Versatile Formal Analyzer of Neural Networks}
\titlerunning{Marabou 2.0}
\author{
Haoze Wu \inst{1} \and 
Omri Isac\inst{2} \and 
Aleksandar Zelji\'c\inst{1} \and 
Teruhiro Tagomori\inst{1,3} \and \\
Matthew Daggitt\inst{4} \and 
Wen Kokke\inst{5} \and 
Idan Refaeli\inst{2} \and 
Guy Amir\inst{2} \and 
Kyle Julian\inst{1} \and \\
Shahaf Bassan\inst{2} \and
Pei Huang\inst{1}\and
Ori Lahav\inst{2} \and 
Min Wu\inst{1} \and 
Min Zhang\inst{6} \and \\
Ekaterina Komendantskaya\inst{4} \and
Guy Katz\inst{2} \and 
Clark Barrett\inst{1}
}

\authorrunning{H. Wu et al.}
%First names are abbreviated in the running head.
%If there are more than two authors, 'et al.' is used.

\institute{
Stanford University, USA \\ 
\and The Hebrew University of Jerusalem, Israel \\
\and NRI Secure \\
\and Heriot-Watt University, UK\\
\and University of Strathclyde, UK \\
%\and The University of Western Australia, Australia \\
\and East China Normal University, China \\
}

\maketitle              % typeset the header of the contribution
\begin{abstract}
This paper serves as a comprehensive system description of 
version 2.0 of the \marabou framework for formal analysis of 
neural networks. We discuss the tool's architectural design and 
highlight the major features and components introduced
since its initial release.
\end{abstract}

\section{Introduction}

With the increasing pervasiveness of deep neural networks (DNNs), the
formal analysis of DNNs has become a burgeoning research field within
the formal methods community. Multiple DNN reasoners have been
proposed in the past few years, including
$\alpha$-$\beta$-CROWN~\cite{zhang2022general,wang2021beta,xu2020automatic}, 
ERAN~\cite{kpoly,deeppoly,singh2019boosting}, 
\marabou~\cite{katz2019marabou}, 
MN-BaB~\cite{ferrari2022complete}, 
NNV~\cite{tran2020nnv,nnvTwo}, nnenum~\cite{nnenum}, VeriNet~\cite{verinet,deepsplit}, 
and many others. 
%These tools have been used in verifying numerous properties of interest in
%DNNs, such as robustness~\cite{AI2}, safety~\cite{julian2019guaranteeing}, 
%and fairness~\cite{khedr2023certifair}.
%in application domains such as perception~\cite{}, systems~\cite{}, robotics~\cite{}, etc. 
%In addition, formal analysis has seen application in explainable
%AI~\cite{ignatiev2019abduction}, 
%neural network pruning~\cite{}, and ensemble model
%selection~\cite{}.

We focus here on the \marabou~\cite{katz2019marabou} tool, which has
been used by the research community in a wide range of formal
DNN reasoning applications (e.g.,
\cite{guidotti2020verification,christakis2021automated,la2021a,funk2021learning,
sun2022vpn,xie2022neuro,bauer2022specrepair,geng2023towards,vinzent2023neural,
yerushalmi2023enhancing,huang2023robustness,liu2023verifying},
inter alia). Initially, \marabou was introduced as a from-scratch
re-implementation of the Reluplex~\cite{katz2017reluplex} decision
procedure, with a native linear programming engine and limited support
for DNN-level reasoning. Over the years, fundamental changes have
been made to the tool, not only from an algorithmic perspective but also
to its engineering and implementation. 

This paper introduces version 2.0 of \marabou.
Compared to its predecessor, \marabouTwo:
\begin{inparaenum}[(i)]
\item employs a new build/test system;
\item has an optimized core system architecture;
\item runs an improved decision procedure and abstract
  interpretation techniques;
\item handles a wider range of activation functions;
\item supports proof
  production;
\item supports additional input formats; and
\item contains a more powerful Python API.
\end{inparaenum}
Due to these changes, the original system
description~\cite{katz2019marabou} no longer gives an accurate
account of the tool's current capabilities. Our
goal in this paper is to close this gap and provide a
comprehensive description of the current \marabou system.
We highlight the major features introduced since the initial version,
describe a few of its many recent uses, and report on 
its performance, as demonstrated by the VNN-COMP'23 results 
and additional runtime comparisons against an early version 
of \marabou.

%%% Local Variables:
%%% mode: latex
%%% TeX-master: "main"
%%% End:

\section{Architecture and Core Components}\label{sec:sys}

\begin{figure}[t]
\centering
\begin{adjustbox}{width=1.02\textwidth,center}
\tikzset{every picture/.style={line width=0.75pt}} %set default line width to 0.75pt        

\begin{tikzpicture}[x=0.75pt,y=0.75pt,yscale=-1,xscale=1]
%uncomment if require: \path (0,304); %set diagram left start at 0, and has height of 304

%Shape: Rectangle [id:dp8016354261263359] 
\draw  [fill={rgb, 255:red, 226; green, 241; blue, 255 }  ,fill opacity=1 ] (285.6,11) -- (650,11) -- (650,299) -- (285.6,299) -- cycle ;
%Shape: Rectangle [id:dp02784304133550375] 
\draw  [fill={rgb, 255:red, 200; green, 229; blue, 255 }  ,fill opacity=1 ] (295,62) -- (640,62) -- (640,231) -- (295,231) -- cycle ;
%Shape: Rectangle [id:dp8743933540810724] 
\draw  [color={rgb, 255:red, 0; green, 0; blue, 0 }  ,draw opacity=1 ][fill={rgb, 255:red, 246; green, 246; blue, 246 }  ,fill opacity=1 ] (85,147.09) -- (263.5,147.09) -- (263.5,262.59) -- (85,262.59) -- cycle ;
%Shape: Rectangle [id:dp028577701411090617] 
\draw  [fill={rgb, 255:red, 226; green, 241; blue, 255 }  ,fill opacity=1 ] (81,38) -- (268,38) -- (268,117) -- (81,117) -- cycle ;
%Shape: Rectangle [id:dp4447271047593748] 
\draw  [fill={rgb, 255:red, 200; green, 229; blue, 255 }  ,fill opacity=1 ] (91.58,68) -- (185,68) -- (185,88.5) -- (91.58,88.5) -- cycle ;
%Shape: Rectangle [id:dp28352537663222543] 
\draw  [fill={rgb, 255:red, 200; green, 229; blue, 255 }  ,fill opacity=1 ] (185,68) -- (257.02,68) -- (257.02,88.5) -- (185,88.5) -- cycle ;
%Shape: Rectangle [id:dp9397020087870935] 
\draw  [fill={rgb, 255:red, 200; green, 229; blue, 255 }  ,fill opacity=1 ] (91.58,88.5) -- (257.02,88.5) -- (257.02,108) -- (91.58,108) -- cycle ;
%Shape: Rectangle [id:dp5631603189654832] 
\draw  [fill={rgb, 255:red, 229; green, 229; blue, 229 }  ,fill opacity=1 ] (112,176) -- (238,176) -- (238,252.5) -- (112,252.5) -- cycle ;
%Shape: Rectangle [id:dp636251182459789] 
\draw  [fill={rgb, 255:red, 229; green, 229; blue, 229 }  ,fill opacity=1 ] (93.72,195.19) -- (254,195.19) -- (254,214.5) -- (93.72,214.5) -- cycle ;
%Shape: Rectangle [id:dp20210456916382635] 
\draw  [fill={rgb, 255:red, 168; green, 205; blue, 243 }  ,fill opacity=1 ] (471.8,91.3) -- (631,91.3) -- (631,193) -- (471.8,193) -- cycle ;
%Shape: Rectangle [id:dp7659278313220421] 
\draw  [fill={rgb, 255:red, 132; green, 184; blue, 238 }  ,fill opacity=1 ] (479,141.65) -- (624,141.65) -- (624,187) -- (479,187) -- cycle ;
%Shape: Rectangle [id:dp5869620302358352] 
\draw  [fill={rgb, 255:red, 168; green, 205; blue, 243 }  ,fill opacity=1 ] (471.8,201) -- (631,201) -- (631,222.19) -- (471.8,222.19) -- cycle ;
%Shape: Rectangle [id:dp22656850904003312] 
\draw  [fill={rgb, 255:red, 132; green, 184; blue, 238 }  ,fill opacity=1 ] (479,113.34) -- (624,113.34) -- (624,136) -- (479,136) -- cycle ;
%Shape: Rectangle [id:dp4284174153287219] 
\draw  [fill={rgb, 255:red, 77; green, 154; blue, 234 }  ,fill opacity=1 ] (485,162) -- (617,162) -- (617,181) -- (485,181) -- cycle ;
%Shape: Rectangle [id:dp9339422847740189] 
\draw  [fill={rgb, 255:red, 200; green, 229; blue, 255 }  ,fill opacity=1 ] (295,43) -- (429,43) -- (429,62) -- (295,62) -- cycle ;
%Shape: Rectangle [id:dp6221869131512794] 
\draw  [fill={rgb, 255:red, 200; green, 229; blue, 255 }  ,fill opacity=1 ] (429,43) -- (530,43) -- (530,62) -- (429,62) -- cycle ;
%Shape: Rectangle [id:dp055614768726739205] 
\draw  [fill={rgb, 255:red, 168; green, 205; blue, 243 }  ,fill opacity=1 ] (304.05,161) -- (463.25,161) -- (463.25,222.19) -- (304.05,222.19) -- cycle ;
%Shape: Rectangle [id:dp8050928560000497] 
\draw  [fill={rgb, 255:red, 132; green, 184; blue, 238 }  ,fill opacity=1 ] (310.3,188.59) -- (457,188.59) -- (457,215) -- (310.3,215) -- cycle ;
%Straight Lines [id:da08391247194502527] 
\draw    (173.45,117.5) -- (173.45,144.5) ;
\draw [shift={(173.45,146.5)}, rotate = 270] [fill={rgb, 255:red, 0; green, 0; blue, 0 }  ][line width=0.08]  [draw opacity=0] (12,-3) -- (0,0) -- (12,3) -- cycle    ;
%Shape: Rectangle [id:dp6619711453518295] 
\draw  [fill={rgb, 255:red, 229; green, 229; blue, 229 }  ,fill opacity=1 ] (93.72,214.5) -- (254,214.5) -- (254,233.81) -- (93.72,233.81) -- cycle ;
%Shape: Rectangle [id:dp511399775232342] 
\draw  [fill={rgb, 255:red, 229; green, 229; blue, 229 }  ,fill opacity=1 ] (93.72,233.81) -- (254,233.81) -- (254,253.12) -- (93.72,253.12) -- cycle ;
%Shape: Rectangle [id:dp3061824319811828] 
\draw  [fill={rgb, 255:red, 229; green, 229; blue, 229 }  ,fill opacity=1 ] (93.72,175.88) -- (254,175.88) -- (254,195.19) -- (93.72,195.19) -- cycle ;
%Shape: Rectangle [id:dp4854962456811158] 
\draw  [fill={rgb, 255:red, 168; green, 205; blue, 243 }  ,fill opacity=1 ] (304.05,91.3) -- (463.25,91.3) -- (463.25,153) -- (304.05,153) -- cycle ;
%Shape: Rectangle [id:dp9302428793398512] 
\draw  [fill={rgb, 255:red, 200; green, 229; blue, 255 }  ,fill opacity=1 ] (295,239) -- (463.25,239) -- (463.25,287.7) -- (295,287.7) -- cycle ;
%Shape: Rectangle [id:dp9688175220093953] 
\draw  [fill={rgb, 255:red, 200; green, 229; blue, 255 }  ,fill opacity=1 ] (471.75,239) -- (640,239) -- (640,287.7) -- (471.75,287.7) -- cycle ;
%Straight Lines [id:da6868493230819666] 
\draw    (264,200) -- (283.8,160) ;
\draw [shift={(285,158)}, rotate = 117] [fill={rgb, 255:red, 0; green, 0; blue, 0 }  ][line width=0.08]  [draw opacity=0] (12,-3) -- (0,0) -- (12,3) -- cycle    ;

% Text Node
\draw (174.3,98.25) node  [font=\small] [align=left] {C++ API};
% Text Node
\draw (215.47,78.25) node  [font=\small] [align=left] {CLI};
% Text Node
\draw (138.79,78.25) node  [font=\small] [align=left] {Python API};
% Text Node
\draw (479.5,52.5) node  [font=\fontsize{8pt}{8pt}\selectfont] [align=left] {CEGAR Solver};
% Text Node
\draw (362,52.5) node  [font=\fontsize{8pt}{8pt}\selectfont] [align=left] {Multi-thread Manager};
% Text Node
\draw (383.65,201.8) node  [font=\fontsize{8pt}{8pt}\selectfont] [align=left] {Abstract Interpretations};
% Text Node
\draw (551,171.5) node  [font=\fontsize{7.8pt}{7.8pt}\selectfont] [align=left]{Simplex Engine};
% Text Node
\draw (551.4,103.15) node  [font=\small] [align=left]{SMT Solver};
% Text Node
\draw (551.4,211.59) node  [font=\small] [align=left] 
{(MI)LP Interface};
% Text Node
\draw (551.5,151.5) node  [font=\fontsize{8pt}{8pt}\selectfont] [align=left] 
{Linear Solving Engine};
% Text Node
\draw (551.5,124.67) node  [font=\fontsize{8pt}{8pt}\selectfont] [align=left] 
{SMT Core};
% Text Node
\draw (555.88,263.35) node  [font=\normalsize] [align=center] 
{Context-dependent\\Data Structures};
% Text Node
\draw (383.4,173.85) node  [font=\small] [align=left] 
{Network-level Reasoner};
% Text Node
\draw (383.65,122.15) node  [font=\small] [align=left] {Preprocessor};
% Text Node
\draw (379.13,263.35) node  [font=\normalsize] [align=left] {Proof Module};
% Text Node
\draw (467.5,28) node  [font=\large] [align=left] {Back End};
% Text Node
\draw (467.5,76) node  [font=\normalsize] [align=left] {Engine};
% Text Node
\draw (175,243.47) node  [font=\fontsize{8pt}{8pt}\selectfont] [align=left] {Non-linear Constraints};
% Text Node
\draw (173.86,204.84) node  [font=\fontsize{8pt}{8pt}\selectfont] [align=left] {(In)equations};
% Text Node
\draw (173.86,224.16) node  [font=\fontsize{8pt}{8pt}\selectfont] [align=left] {Piecewise-linear Constraints};
% Text Node
\draw (175,185.53) node  [font=\fontsize{8pt}{8pt}\selectfont] [align=left] {Bounds};
% Text Node
\draw (173.86,161.84) node  [font=\normalsize] [align=left] 
{Input Query};
% Text Node
\draw (173.91,53) node  [font=\large] [align=left] 
{Front End};

\end{tikzpicture}
\end{adjustbox}
\caption{High-level overview of \marabouTwo's system architecture.}
\label{fig:sys}
\end{figure}
%\omri{Currently we do not describe the SMTLIB writer module. Should we add the description, or modify the figure (and add 
% the Checker module instead)?}
% \andrew{Will modify the figure.}

In this section, we discuss the core components of \marabouTwo. 
An overview of its system architecture is given in Figure~\ref{fig:sys}. 
At a high level, \marabou performs satisfiability checking on a set of 
linear and non-linear constraints, supplied through one of the front-end 
interfaces. The constraints typically represent a verification query 
over a neural network and are stored in an \cls{InputQuery} object. 
We distinguish variable bounds from other linear constraints, and 
piecewise-linear constraints (which can be reduced to linear 
constraints via case analysis) from more general, non-linear constraints. 

Variables are represented as consecutive indices starting from 0. 
(In)equations are represented as \cls{Equation} objects. Piecewise-linear 
constraints are represented by objects of classes that inherit from the 
\cls{PiecewiseLinearConstraint} abstract class. The abstract class defines 
the key interface methods that are implemented in each sub-class. This way, 
all piecewise-linear constraints are handled uniformly in the back end. 
Similarly, each other type of non-linear constraint is implemented as a sub-class 
of the new \cls{NonlinearConstraint} abstract class. 
Initially, \marabou only supported the ReLU and Max constraints. 
In \marabouTwo, over ten types of non-linear constraints 
(listed in \extref{Appendix~\ref{subapp:supportedconstraints}}) 
are supported.

%\katya{This is a significnt extension, does it make sense to put a comma and breifly say something like "an extension that was possible thanks to ..." }
%\andrew{Hum.. Not sure what to say here. Could you be more specific?}

\subsection{Engine}

The centerpiece of \marabou is called the Engine, which reasons about the 
satisfiability of the input query. The engine consists of several components: 
the Preprocessor, which performs rewrites and simplifications; the Network-level 
Reasoner, which maintains the network architecture and performs all 
analyses that require this knowledge; the SMT Solver, which houses complete 
decision procedures for sets of linear and piecewise-linear constraints; 
and the (MI)LP Interface, which manages interactions with external (MI)LP solvers 
for certain optional solving modes as explained below. 
%\katya{I would add "e.g." after certain solving modes and mention what they are}
%\andrew{I left it vague to prevent diverging the readers, since those modes are all optional. I rewrote a little.}

Two additional modules are built on top of the Engine. The Multi-thread Manager 
spawns multiple Engine instances to take advantage of multiple processors. The 
CEGAR Solver performs incremental linearization~\cite{wu2023toward,cimatti2018incremental} 
%\katya{citation [60] has strange symbols * near author names. Not sure it is intended}\andrew{Thanks. Fixed!}
for non-linear constraints that cannot be precisely handled by the SMT Solver. 
%\katya{Because you do not seem to come back to CEGAR again (and it occupies a side bar in the diagram), it may help to mention the mode of its operation: i.e. anytime non-linear constraints are detected, CEGAR is used, correct? in that sense, it is not an optional part of the pipeline?}
%\andrew{rewritten the CEGAR paragraph a little.}

\subsubsection{Preprocessor.}
Every verification query first goes through multiple preprocessing
passes, which \emph{normalize}, \emph{simplify}, and \emph{rewrite} the query. 
One new normalizing pass introduces auxiliary variables and entailed linear
constraints for each of the piecewise-linear constraints, so that
case splits on the piecewise-linear constraints can be represented as bound updates 
and consequently do not require adding new equations.\footnote{For example, for a piece-wise linear 
constraint $y = \max(x_1, x_2)$, we would introduce 
$c_1: y - x_1 = a_1 \land a_1 \geq 0 \land y - x_2 = a_2 \land a_2 \geq 0$, 
where $a_1$ and $a_2$ are fresh variables. This way, case splits on this constraint 
can be represented as $c_2: a_1 \leq 0$ and $c_3: a_2 \leq 0$, respectively. 
This preprocessing pass preserves satisfiability because the original constraint 
is equisatisfiable to $c_1 \land (c_2 \lor c_3)$.}
This accelerates the underlying Simplex engine, as
explained in the SMT Solver section below.
%\matthew{Maybe give a section reference instead of "later"?}
%\andrew{done}
Another significant preprocessing pass involves
iterative bound propagation over all
constraints. In this process, piecewise linear constraints might
collapse into linear constraints and be removed. This pass was present
in \marabouOne, but could become a runtime bottleneck; 
whereas \marabouTwo employs a data structure
optimization that leads to a  $\sim$60x speed up. Finally, the
preprocessor merges any variables discovered to be equal to each
other and also eliminates any constant variables. This results
in updates to the variable indices, and therefore a mapping from 
old indices to new ones needs to be maintained for retrieving 
satisfying assignments.
%\matthew{"This results in updates *to* the variable indices"?}
%\andrew{fixed. Thanks!}
%\guy{Should we give an example here? E.g., of how a case split is
%  reduced to bound tightening?}
%\andrew{I had an example that was commented out. I added it back as a footnote.}

\subsubsection{SMT Solver.}
The SMT Solver module implements a sound and
complete, lazy-DPLL(T)-based procedure for deciding the satisfiability
of a set of linear and piecewise-linear constraints. It performs case
analysis on the piecewise-linear constraints and, at each search
state, employs a specialized procedure to iteratively search for 
an assignment satisfying both the linear and non-linear constraints.

Presently, the DeepSoI procedure~\cite{wu2022efficient} has
replaced the Reluplex
procedure~\cite{katz2017reluplex,katz2019marabou} as \marabou's default
procedure to run at each search state. The former provably converges
to a satisfying assignment (if it exists) and empirically consistently
outperforms the latter. DeepSoI extends the canonical
sum-of-infeasibilities method in convex
optimization~\cite{boyd2004convex}, which determines the
satisfiability of a set of linear constraints by minimizing a cost
function that represents the total violation of the constraints by the
current assignment. The constraints are satisfiable if and only if the
optimal value is 0. Similarly, DeepSoI formulates a cost function that
represents the total violation of the current piecewise-linear
constraints and uses a convex solver to stochastically minimize the
cost function with respect to the convex relaxation of the current
constraints. In addition, DeepSoI also informs the branching
heuristics of the SMT Core, which performs a case split on the
piecewise-linear constraint with the largest impact (measured by the
\emph{pseudocost} metric~\cite{wu2022efficient}) on the cost
function. The DeepSoI procedure is implemented for all
supported piecewise-linear activation functions. The convex solver 
can be instantiated either with the native Simplex engine 
%(which was extended to perform not just satisfiability-checking but also optimization)
%\guy{Does this part about optimization refer to the work by Chris Strong? I
%  believe it was never merged to master} 
%\andrew{I was referring to the small changes made to the Simplex core here and there to 
%make it take arbitrary cost function instead of just phase I.  I removed this optimization sentence since we need to cut anyway.}  
or with an external LP solver via the (MI)LP interface
(detailed below). The latter can be more efficient but requires
the use of external commercial solvers.

One optimization in \marabouTwo's Simplex engine is that once the tableau
has been initialized, it avoids introducing any new equations --- a costly operation that requires re-computing the tableau from scratch. 
This is achieved by implementing case-splitting and backtracking
as updates on variable bounds (as mentioned above), which only 
requires minimal updates to the tableau state. 
By our measure, this optimization reduces the runtime of the
Simplex engine by over 50\%. Moreover, the memory footprint of the solver
is also drastically decreased, as the SMT Core no longer needs 
to save the entire tableau state during case-splitting (to be restored 
during backtracking). 

\subsubsection{Network-level Reasoner.} 
Over the past few years, numerous papers (e.g., ~\cite{wang2018efficient,deeppoly,zhang2018efficient,prima,zelazny2022reducing}, inter alia) have
proposed abstract interpretation techniques that rely on network-level
reasoning (e.g., propagating the input bounds layer by layer to
tighten output bounds). 
These analyses can be viewed as a stand-alone, incomplete DNN verification 
procedure, or as in-processing bound tightening passes for the SMT Solver. \marabouTwo
features a brand new \cls{NetworkLevelReasoner} class that supports this type
of analysis. The class maintains the neural network topology as a
directed acyclic graph, where each node is a \cls{Layer} object. The
\cls{Layer} class records key information such as weights, source
layers, and mappings between neuron indices and variable
indices. Currently, seven different analyses are implemented:
\begin{inparaenum}{[i]}
\item interval bound propagation~\cite{gowal2018effectiveness};
\item symbolic bound propagation~\cite{wang2018efficient};
\item DeepPoly/CROWN
  analysis~\cite{deeppoly,zhang2018efficient};
\item LP-based bound tightening~\cite{mipverify};
\item Forward-backward analysis~\cite{wu2022scalable};
\item MILP-based bound tightening~\cite{mipverify}; and
\item iterative propagation~\cite{wu2020parallelization}.
\end{inparaenum}
Analyses 2--7 are implemented in a parallelizable manner, and analyses 4--7
require calls to an external LP solver. By default, the DeepPoly/CROWN analysis 
is performed. The Network-level Reasoner tightly interleaves with the SMT Solver: 
the network-level reasoning is executed any time a new search state is reached 
(with the most up-to-date variable bounds), and the derived bound tightenings 
are immediately fed back to the search procedure. 

It is noteworthy that the user does not have to explicitly provide the
neural network topology to enable network-level reasoning.
%\matthew{I'm very curious as to what the advantage of this is. Maybe we could add a sentence?} 
%\andrew{See use case 1 and 3. I thought about adding something but cannot find a single
%sentence that does the job..}
Instead, the network architecture is \emph{automatically inferred} from the given set of 
linear and non-linear constraints, via the \emph{constructNetworkLevelReasoner} 
method in the \cls{InputQuery} class. The Network-level Reasoner is only
initialized if such inference is successful. Apart from the abstract
interpretation passes, the Network-level Reasoner can also evaluate concrete 
inputs. This is used to implement the LP-based bound 
tightening optimization introduced by the NNV tool~\cite{tran2020nnv}. 

\subsubsection{(MI)LP Interface.}

\marabou can now optionally be configured to invoke the Gurobi
Optimizer~\cite{gurobi}, a state-of-the-art Mixed Integer Linear
Programming (MILP) solver. 
The \cls{GurobiWrapper} class contains methods to
construct a MILP problem and invoke the solver. The \cls{MILPEncoder}
class is in charge of encoding the current set of linear and
non-linear constraints as (MI)LP constraints. Piecewise-linear
constraints can either be encoded precisely, or replaced with a
convex relaxation, resulting in a linear program. For
other non-linear constraints, only the latter option is available. The
(MI)LP interface presently has three usages in the code base. Two have
already been mentioned, i.e., in some of the abstract interpretation
passes and optionally in the DeepSoI procedure. Additionally, when
\marabou is compiled with Gurobi, a \texttt{--milp} mode is
available, in which the Engine performs preprocessing and abstract
interpretation passes, and then directly encodes the verification 
problem as a MILP problem to be solved by Gurobi. 
%\matthew{Should be "a --milp" rather than "an --milp"?}\andrew{fixed}  
The mode is motivated by the observation that the performance of Gurobi 
and the SMT Solver can be complementary~\cite{wu2022efficient,strong2023global}.

\subsubsection{Multi-thread Manager.} Parallelization is an important way 
to improve verification efficiency. \marabou supports two modes of 
parallelization, both managed by the new \cls{MultiThreadManager} class: 
the \emph{split-and-conquer} mode~\cite{wu2020parallelization} and the 
\emph{portfolio} mode. In the split-and-conquer mode, the original query is
dynamically partitioned and re-partitioned into independent sub-queries, to 
be handled by idle workers. The partitioning strategy is implemented as a 
sub-class of the \cls{QueryDivider} abstract class. Currently, two strategies 
are available: one partitions the intervals of the input variables; the 
other splits on piecewise linear constraints. By default, 
the former is used only when the input dimension is less than or equal to ten. In the portfolio 
mode, each worker solves the same query with a different random seed, which
takes advantage of the stochastic nature of the DeepSoI procedure. Developing 
an interface to define richer kinds of portfolios is work in progress.

\subsubsection{CEGAR Solver.}
%\katya{Should this text be closer to CEGAR text under the "Engine" section?}
While the DNN verification community has by and large focused on
piecewise-linear activation functions, other classes of non-linear
connections exist and are commonly used for certain architectures~\cite{huang2018multimodal,vaswani2017attention}. 
Apart from introducing support for non-linear constraints in the Preprocessor 
and the Network-level Reasoner, the latest \marabou version also incorporates a counter-example guided 
abstraction refinement (CEGAR) solving mode~\cite{wu2023toward}, 
based on incremental linearization~\cite{cimatti2018incremental} to enable
more precise reasoning about non-linear constraints that are not piecewise linear.
Currently, the CEGAR solver only supports Sigmoid and Tanh, but the
module can be extended to handle other activation functions.

%\marabou now supports the four aforementioned functions in its preprocessing and DeepPoly/CROWN
%abstract interpretation operations.
%\katya{When these two are joined in one sentences, it is hard to read: there is preprocessing and there is a choice of solving algorithm, correct? maybe distinction should be made clear...}
%In addition, a 
%%counter-example guided abstraction refinement (CEGAR) solving mode~\cite{wu2023toward}, 
%based on incremental linearization~\cite{cimatti2018incremental} is also available
%to precisely handle transcendental activation functions. Currently, only Sigmoid and Tanh 
%are supported, but the module can be extended to other activation functions. \katya{I am a bit lost. The para says 4  transcendental functions are supported, and now %it says 2.}
%\katya{ Generally, the picture is unclear.
%So there are two ways to handle non-linearity: preprocessing with CEGAR and then suing SMT-solving or alternatively using CROWN instead of SMT solving. Is this what the paragraph is trying to say?}
%\andrew{I re-wrote the sentence a little} 

%\subsubsection{SMT_LIB Writer}

\subsection{Context-Dependent Data-Structures}

When performing a case split or backtracking to a previous search
state, the SMT Core needs to save or restore information such as
variable bounds and the phase status of each piecewise-linear
constraint (e.g., is a ReLU currently active, inactive, or
unfixed). To efficiently support these operations, \marabouTwo 
uses the notion of a context level (borrowed from the CVC4 SMT
solver~\cite{barrett2011cvc4}), and stores the aforementioned
information in \emph{context-dependent data structures}. These data
structures behave similarly to their standard counterparts, except
that they are associated with a context level and \emph{automatically}
save and restore their state as the context increases or decreases. % While this refactoring has by far led to little runtime improvement and mild reduction in memory footprint, it has significantly reduced the implementation overhead when it comes to saving and restoring solver states (one could simply push and pop the context level)
This major refactoring has greatly simplified the implementation of saving and restoring solver states and is an important milestone in an ongoing effort to integrate
a full-blown Conflict-Driven Clause-Learning (CDCL) mechanism into \marabou.

\subsection{Proof Module}

A proof module has recently been 
introduced into \marabou, enabling it to optionally produce proof certificates after an unsatisfiable (\unsat)~\cite{IsBaZhKa22} result.
This is common practice in the SAT and SMT communities and is aimed at
ensuring solver reliability.
\marabou produces proof certificates based on a constructive variant of the Farkas lemma~\cite{Va98}, which ensures the existence of a \emph{proof
vector} that witnesses the unsatisfiability of a linear program. Specifically, the \emph{proof vector} corresponds to a linear equation that is violated by the variable bounds~\cite{IsBaZhKa22}. 
The full certificate of \unsat is comprised of a \emph{proof tree},
whose nodes represent the search states explored during the solving.
 Each node may contain a list of \emph{lemmas} %\omri{Leaves may contain lemmas too} 
that are used as additional constraints in its descendent
nodes; and each leaf node contains the proof vector for the
unsatisfiability of the corresponding sub-query. The lemmas
encapsulate some of the variable bounds, newly derived by the
piecewiese-linear constraints of the query, and require their own
witnesses (i.e., proof vectors).
The \cls{BoundExplainer} class is
responsible for constructing all proof vectors, for updating them
during execution, and for appending them to the node. The proof tree itself is implemented using the \cls{UnsatCertificateNode} class.

When the solver is run in proof-production mode, the Proof module
closely tracks the steps of the SMT Solver module and constructs the
proof tree on the fly: new nodes are added to the tree whenever a case
split is performed; and a new proof vector is generated whenever a
lemma is learned or \unsat is derived for a sub-query. If the Engine
concludes that the entire query is \unsat, a proof checker
(implemented as an instance of the \cls{Checker} class) will be
triggered to certify the proof tree. It does so by traversing the tree
and certifying the correctness of the lemmas and the unsatisfiability of the leaf nodes. A formally verified and precise proof-checker is currently under development~\cite{DeIsPaStKoKa23}.  Note that, currently, proof production mode is only compatible with %numerous piecewise-linear activation functions ($\relu$, $\sign$, $maxpool$ and absolute-value), though it is only compatible with 
a subset of the features supported by \marabou. Adding support for the remaining features (e.g., for the parallel solving mode) is an ongoing endeavor.

\subsection{Front End}

\marabou provides interfaces to prepare input queries and invoke the back-end solver in multiple 
ways. The \marabou executable can be run on the command line, taking in network/property/query files 
in supported formats. The Python and C++ APIs support this functionality as well, but also 
contain methods to add arbitrary linear and (supported) non-linear constraints. 
In addition, a layer on top of the Python API was added to \marabouTwo which
allows users to define constraints in a more \emph{Pythonic} manner, resulting in 
more succinct code. For example, suppose one wants to check whether the first output of a network 
(stored in the ONNX format) can be less than or equal to half of its second output, when the 
first input is greater than or equal to $0.1$. Figure~\ref{fig:base} shows how to perform this check 
with the base Python API, while Figure~\ref{fig:pythonic} exhibits the ``Pythonic'' API.

\begin{figure}[t]
\centering
\begin{subfigure}{0.48\linewidth}
\begin{scriptsize}
\begin{verbatim}
Q = Marabou.read_onnx("model.onnx")
X, Y = Q.inputVars[0], Q.outputVars[0]
Q.setLowerBound(X[0], 0.1)
Q.addInequality([Y[0], Y[1]], [1, -0.5], 0)
Q.solve()
\end{verbatim}
\end{scriptsize}
\caption{The base Python API}\label{fig:base}
\end{subfigure}
%\quad
\begin{subfigure}{0.48\linewidth}        
\begin{scriptsize}
\begin{verbatim}
Q = Marabou.read_onnx("model.onnx")
X, Y = Q.inputVars[0], Q.outputVars[0]
Q.addConstraint(Var(X[0]) >= 0.1)
Q.addConstraint(Var(Y[0]) <= 0.5 * Var(Y[1]))
Q.solve()
\end{verbatim}
\end{scriptsize}
\caption{The ``Pythonic'' API}\label{fig:pythonic}
\end{subfigure}
\label{pythonic}
\caption{Two ways to define the same verification query through the Python API.}
\end{figure}

Typically, a query consists of the encoding of (one or several) neural
networks and the encoding of a property on the network(s). To encode a
neural network, the user has two options: 1) pass in a neural network
file to be parsed by one of the neural network parsers; or 2) manually add constraints to encode
the neural network. The main network format for \marabouTwo is now ONNX, towards which the neural 
network verification community is converging. The NNet format and the Tensorflow protobuf format 
are still supported but will likely be phased out in the long run.
To encode the property on top of the neural network encoding, the user can 1) pass in a property file to be parsed by one of the property parsers;
or 2) manually encode the property. Currently \marabou has two property parsers, one for a native property file format~\cite{katz2019marabou}, and a new one for the VNN-LIB format, supporting the standardization effort of the community.
%\katya{is item 1) related to the last sentence about VNNLIB, or is it some different parser, and VNNLIN parser is given in addition? It is not entirely clear}
%\andrew{Yes, it is. I made this clearer.}

In addition to the aforementioned network and property file formats, \marabou also supports a native query file format that describes a set of linear and non-linear constraints. This can be dumped/parsed from all interfaces.

\subsection{Availability, License, and Installation}
\marabou is available under the permissive modified BSD open-source
license, and runs on Linux and macOS machines. 
The tool can be built from scratch using CMake. \marabou is now also 
available on The Python Package Index (PyPI) and can be installed 
through \software{pip}. The latest version of Marabou is available at: \url{https://github.com/NeuralNetworkVerification/Marabou}. The artifact associated with this tool description is archived on Zenodo~\cite{artifact}.

\section{Highlighted Features and Applications}\label{sec:casestudies}

In terms of performance, \marabou is on par with state-of-the-art verification tools. 
In the latest VNN-COMP~\cite{brix2023fourth}, \marabou won the second place overall, 
and scored the highest among all CPU-based verifiers. We summarize the main results in 
\extref{Appendix~\ref{app:vnn-comp}}.
%\katya{Section~ B $\rightarrow$ Appendix B.}\guy{I think we
%   said we would borrow a table from the VNN comp report. If there's
%   no space, maybe we can put it in the appendix, and mention it
%   here?} \andrew{Will put it in the appendix}
In this section, we focus on the usability aspect of \marabou, and 
highlight some of its recent applications --- as well as the features that make
them possible. We believe this diverse set of use cases (as well as the relevant scripts 
in the artifact~\cite{artifact}) serve as valuable examples, which will inspire new 
ways to apply the solver. 
\ifnotarxiv
More use cases can be found in \extref{Appendix~\ref{app:casestudies}}.
\fi
A runtime evaluation of \marabouTwo against an early version appears in Section~\ref{sec:eval}.

\subsubsection{Verifying the Decima job scheduler.}
Recently, Graph Neural Networks (GNNs) have been used to schedule
jobs over multi-user, distributed-computing clusters, achieving state-of-the-art 
job completion time~\cite{decima}. However, concerns remain over whether GNN-based 
solutions satisfy expected cost-critical properties beyond performance. 
\marabou has been used to verify a well-known fairness 
property called \emph{strategy-proofness}~\cite{wu2022scalable} 
for a high-profile, state-of-the-art GNN-based scheduler called Decima~\cite{decima}. 
The verified property states that ``a user cannot get their job scheduled 
earlier by misrepresenting their resource requirement.''
While it is challenging to represent a GNN directly in ONNX~\cite{gitissue}, Marabou's Python API makes it possible to manually encode Decima and the 
specification as a set of linear and non-linear constraints. From these constraints, 
the Network-level  Reasoner
%\guy{The capitalization of 'Network-level
%  Reasoner' should be consistent across the paper}
%\andrew{made a pass to make sure this is the case}
is able to automatically infer a feed-forward structure 
with residual connections and then use it for the purpose of abstract interpretation. 
Notably, \marabou was able to handle the \emph{original} Decima architecture, proving that 
the property holds on the vast majority of the examined job profiles 
but can indeed be violated in some cases.

\subsubsection{Formal XAI.}
Despite their prevalence, DNNs are considered ``black boxes'',
uninterpretable to humans. \emph{Explainable AI} (XAI) aims to understand
DNN decisions to enhance trust. Most XAI methods 
are heuristic-based and 
lack formal correctness
guarantees~\cite{ribeiro2016should, lundberg2017unified,
  ribeiro2018anchors}, which can be problematic for critical,
regulation-heavy systems.
Recent work showed that \marabou can
be utilized as a sub-routine in procedures designed for producing 
\emph{formal and provable} explanations for
DNNs~\cite{wu2022verix, BaAmCoReKa23, bassan2023towards,
  la2021a, huang2023robustness}. For instance, it can be used in
constructing formal \emph{abductive
  explanations}~\cite{ignatiev2019abduction, bassan2023towards}, which
are subsets of input features that are, by themselves, provably sufficient
for determining the DNN's output. This approach has been successfully
applied to large DNNs in the domains of computer
vision~\cite{wu2022verix, bassan2023towards}, NLP~\cite{la2021a}, and
DRL robotic navigation~\cite{BaAmCoReKa23}. These studies
highlight the potential of \marabou in tasks that go beyond formal 
verification.

\subsubsection{Analyzing learning-based robotic systems.}
Deep Reinforcement Learning has extensive application in robotic planning 
and control.
%\marabou is also geared for verifying various real-world reactive systems, 
%due to its ability to encode constraints among subsequent steps. This, in turn, allows efficient and scalable verification of 
%Deep Reinforcement Learning (DRL) controllers, and their various applications, most notably ---  robotic planning and 
%control~\cite{ShWoDh17, HaGuSi16, ZhJoBr20}. 
\marabou has been applied in these settings to analyze different 
safety and liveness properties~\cite{ElKaKaSc21, AmScKa21,AmCoYeMaHaFaKa23,vinzent2023neural}.
For example, Amir et al.~\cite{AmCoYeMaHaFaKa23} used \marabou to detect infinite loops in 
a real-world robotic navigation platform. %In this setting, the control policy 
%takes in the current state and outputs an action that drives an 
%autonomous robot to a target location. 
This was achieved by querying whether there exists a state to which
the robot will always return within a finite number of steps $k$, 
effectively entering an infinite loop.
% an infinite loop.
% such that starting from that state, the neural network control policy will drive the robot back
% to that state in  $k$ steps ($k$ is a constant)? 
A multi-step property like this can be conveniently encoded in \marabou, by 
\begin{inparaenum}[(i)]
\item 
encoding $k$ copies of the control policy; 
\item 
for each time-step $t$, encoding the system transition as constraints over
the current state (input to the policy at $t$), 
the decided action (output of the policy at $t$), and the next state 
(input to the policy at $t+1$); and
\item 
encoding the ``loop'' constraint that 
the initial state ($t_{1}$) is equal to the final state ($t_k$). 
\end{inparaenum}
From this set of constraints, the Network-level Reasoner can infer the structure 
of and perform abstract interpretations over a \emph{concatenated network}, 
where the input is the initial state and the output is the final state. 
Moreover, due to the low input dimension, the 
split-and-conquer mode in the Multi-thread Manager can be used to perform input-splitting, 
effectively searching for such loops in independent input regions in parallel.
% This configration allows the user to easily encode relations  
%
Notably, \marabou can detect loops in the system for agents 
trained using state-of-the-art RL algorithms,
in cases where gradient/optimization-based approaches fail to find any. 
Loops detected this way have also been observed in the real world~\cite{RobotYoutubeVideo}.

\subsubsection{Proof production for the ACAS-Xu benchmarks.}
A well-studied set of benchmarks in DNN verification derives from an implementation of the
ACAS-Xu airborne system for collision
avoidance~\cite{JuKoOw19}.
%Producing indpendent certificates for the correctness of such safety-critical systems is highly desirable. 
Using \marabou, we were able to
produce certificates of unsatisfiability for these benchmarks for the first time.
\marabou was able to produce certificates for 
113 out of the 180 tested benchmarks, with only mild overhead incurred
by proof generation and certification. 
The proof certificates contained over 1.46 million proof-tree leaves, of which more than $99.99\%$ were certified by the native proof checker, while the remaining were certified by a trusted SMT solver.
% Concretely,
% \marabou was able to successfully produce proof certificates for 
% 113 out of the 180 tested benchmarks, 
% with mild overhead in proof generation and certification.
% \andrew{@Omri, I modified a little. Still not too happy with the sentence though..}
%over $99.99\%$ of the subqueries within these 113 
%benchmarks. 113 were \unsat{}, 
%and \marabou succeeded in producing proof certificates for over 
%The results indicate that proof production causes only a minor overhead, 
%both in proof generation time and proof certification time.
%Furthermore, the proofs
%produced by \marabouTwo where found correct by its checker in almost
%all cases. An external SMT solver was then able to certify the
%correctness of the few remaining proofs.
Additional details are provided in~\cite{IsBaZhKa22}.

\subsubsection{Specifications on neural activation patterns.}

Properties of hidden neurons garner increasing interest~\cite{yosinski2015understanding}, as they shed light
on the internal decision-making process of the neural network. Gopinath et al.~\cite{gopinath2019property}
observed that for a fixed neural network, certain \emph{neuron activation patterns} (NAPs)
%, i.e., on/off statuses of a subset of neurons
empirically entail a fixed prediction. More recently, Geng et al.~\cite{geng2023towards} 
formally verified (using \marabou) the aforementioned property, along with a variety of 
other properties related to NAPs. Specifications related to NAPs can be conveniently encoded 
in \marabou. For example, specifying that a certain ReLU is activated amounts to setting the 
lower bound of the variable corresponding to the ReLU input to 0, using the general 
constraint-encoding methods in the Python/C++ API. Constraints on 
internal neurons, as with other constraints, can be propagated by the 
Preprocessor and Network-level Reasoner to tighten bounds.

%\guy{Wasn't this kind of trick
%  proposed by Corina and Divya Gopinath a few years earlier? In their
%  compositional reasoning work? }%https://arxiv.org/pdf/1904.13215.pdf} 
%\andrew{I was aware of this work. My understanding is that they didn't do any formal verification in the paper but they did try. I re-incorporated their work in this %paragraph.}

\subsubsection{Robustness against semantically meaningful perturbations.}

Considering specifications of perception networks, there is an ongoing
effort in the verification community to go beyond \emph{adversarial
  robustness}~\cite{balunovic2019certifying,katz2022verification,mirman2021robustness,wu2023toward,mohapatra2020towards}.
\marabou has been used to verify robustness against semantically
meaningful perturbations that can be analytically defined/abstracted
as linear constraints on the neural network inputs (e.g., brightness,
uniform haze)~\cite{paterson2021deepcert}.  More recently, \marabou
has also been successfully applied in a neural symbolic approach,
where the correct network behavior is defined with respect to that of
another network~\cite{xie2022neuro,wu2023toward}.  For example, Wu et
al.~\cite{wu2023toward} considered the specification that an image
classifier's prediction does not change with respect to outputs of an
image generative model trained to capture a complex distribution shift
(e.g., change in weather condition). A property like this can be
conveniently defined in \marabou by loading the classifier and the
generator through the Python API and adding the relevant constraints
on/between their input and output variables.

\ifarxiv

\subsubsection{Minimal modifications of neural networks.}
Recently, multiple papers have proposed techniques to \emph{repair} a neural 
network (e.g., by altering selected weights or adding patches) against input
points that result in buggy behaviors. \marabou has also been applied in this
context to obtain provably \emph{minimal} modification to the DNN's weights so 
that a previously mispredicted input would become correctly predicted. 
We refer the reader  to~\cite{refaeli2022minimal, goldberger2020minimal} 
for additional details.

\subsubsection{Verification of quantized neural Networks.}
\emph{Quantization}~\cite{JacobKCZTHAK18} has become a standard technique for deploying deep neural networks on resource-constrained hardware. The process involves replacing floating point arithmetic with integer arithmetic. Recently, Huang, et.al.~\cite{huang2023towards} have leveraged \marabou to verify quantized neural networks (QNNs). While mainstream platforms like PyTorch do not support saving QNN models in the standard ONNX format, \marabou's flexible Python API makes it possible to manually encode the quantization operations with constraints such as Round and Clip.

\subsubsection{Pruning and slicing neural networks.}
Another well-studied approach to reduce the computational cost of deep neural networks is \emph{pruning}~\cite{HanMD15,IaHaMoAsDaKe16}, which involves identifying and removing neurons that contribute minimally to the network's output. Most existing approaches are heuristic-based and lack formal guarantees that 
the simplified network preserves the behavior of the original. In contrast, it has been shown that \marabou can be used as a sub-procedure in a pruning algorithm~\cite{GoFeMaBaKa20,LaKa21}, which is able to reduce the size of DNNs while provably preserving the output of the neural network up to a given error margin. 

\subsubsection{Integrating \marabou into complex system verification.}
There is a growing demand to embed neural network verification into verification of complex systems that rely on neural networks for perception or decision-making~\cite {PasareanuMGYICY23}. There are multiple challenges in verifying such systems, and we mention  two of them here:  (i) currently available provers struggle to overcome the \emph{embedding gap}, which arises because complex system  properties are stated using the abstractions of the problem domain, whereas neural network propeties are stated at a low-level, in terms of normalized real vectors; and (ii) more generally, higher-order theorem provers like HOL, Coq, and Agda can prove more abstract properties of complex systems, but fail at neural network verification; and vice versa: neural network verifiers are not designed to reason about complex hybrid systems. The domain-specific language \emph{Vehicle}~\cite{daggitt2024vehicle,FoMLAS2023:Vehicle_Tutorial_Neural_Network} aims at bridging this gap; and includes \marabou as its default verifier. This choice was made thanks to \marabou's syntax, which is richer and more flexible than available in other, more specialized neural network verifiers.
\fi

%Although much research has been put into identifying bugs in DNNs
%(e.g., through testing or verification), the question of how to act when
%a bug is discovered has received lesser attention. Ideally, we would
%like to alter (``patch'') the network, so that the discovered bug is removed. However, assuming that the network is generally correct, we
%would like this patch to have a small and local effect, targeting only
%the bug in question.

%In one case-study, we used \marabou as the cornerstone of an algorithm
%that, given a DNN and a bug-triggering input $\mathbf{v}$, would find a \emph{minimal}
%modification to the DNN's weights that causes $\mathbf{v}$ to become
%correctly classified.  The minimality of the modification causes the
%network's behavior to remain unchanged for most other, correctly
%classified inputs. 
%%The algorithm is designed to reduce the minimal modification problem into a verification problem, and then dispatch it using \marabou. 
%The assignment obtained from \marabou for that verification query indicates the sought-after minimal modification. 
%We refer the reader  to~\cite{refaeli2022minimal, goldberger2020minimal} for additional details.

%%% Local Variables:
%%% mode: latex
%%% TeX-master: "main"
%%% End:

\section{Runtime Evaluation}\label{sec:eval}

\begin{wrapfigure}{r}{0.55\textwidth}
\ifnotarxiv
\vspace{-1cm}
\fi
    \centering
    \includegraphics[width=0.55\textwidth]{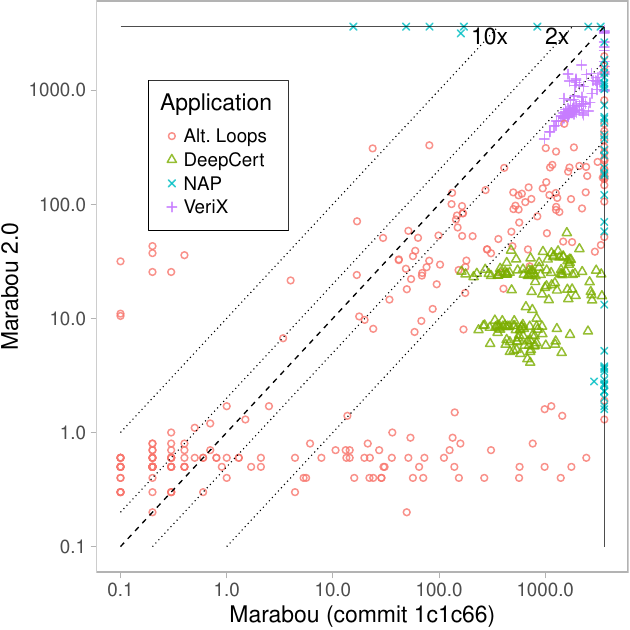}
    \vspace{-6.5mm}
    \caption{Runtime performance of \marabouTwo and an early version of \marabou on four applications supported by both versions.}
    \label{fig:scatter}
\vspace{-0.9cm}
\end{wrapfigure}

We measure the performance improvement in \marabouTwo by
comparing it against an early \marabou version (git commit \url{1c1c66}), 
which can handle ReLU and Max constraints and supports symbolic bound propagation~\cite{wang2018efficient}.
We collected four benchmark sets from the applications described in Section~~\ref{sec:casestudies}: 
Alternating Loop~\cite{AmCoYeMaHaFaKa23}, DeepCert~\cite{paterson2021deepcert}, NAP~\cite{gopinath2019property,geng2023towards},
and VeriX~\cite{wu2022verix}. There are 745 instances in total. 
Details about the benchmarks can be found in \extref{Appendix~\ref{subapp:benchmarks}}.

Figure~\ref{fig:scatter} compares the runtime of the two \marabou versions on all the benchmarks with a 1 hour CPU timeout. 
Each configuration was given 1 core and 8GB of memory. 
Note that \marabouTwo was not configured with external solvers in this experiment. 
We see that \marabouTwo is significantly more efficient for a vast majority of the instances. Upon closer examination, 
an at-least $2\times$ speed-up is achieved on 428 instances and an at-least $10\times$ speed-up is achieved on 263 instances. 
Moreover, \marabouTwo is also significantly more memory efficient, with a median peak usage of 57MB (versus 604MB with the old version).
Solvers'performance on individual benchmarks is reported in \extref{Appendix~\ref{subapp:data}}.

\section{Conclusion and Next Steps}\label{sec:conclusion}

We have summarized the current state of \marabou, a maturing formal analyzer for 
neural-network-enabled systems that is under active development. In its current form, 
\marabou is a versatile and user-friendly toolkit suitable for a wide range of formal analysis tasks. 
Moving forward, we plan to improve \marabou in several dimensions.
Currently, we are actively integrating a CDCL mechanism 
in the SMT Solver module. Given that many applications involve 
repeated invocation of the solver on similar queries, we also plan to support incremental 
solving in the style of pushing and popping constraints, leveraging the newly introduced 
context-dependent data structures. In addition, adding GPU support (in the Network-level Reasoner) 
and handling other types of non-linear constraints are also on the development agenda for \marabou. 

\subsubsection{Acknowledgment} 

The work of Wu, Zelji\'c, Tagomori, Huang and Wu was partially supported by the NSF (grant number 2211505), by the BSF (grant number 2020250), a Ford Alliance Project (199909), the Stanford Center for AI Safety, and the Stanford Institute for Human-Centered Artificial Intelligence (HAI).
The work of Daggit, Kokke and Komendantskaya was partially supported by the EPSRC grant EP/T026952/1, AISEC: AI Secure and Explainable by Construction.
The work of Isac, Refaeli, Amir, Bassan, Lahav and Katz was partially funded by the
ISF (grant number 3420/21), by the BSF (grant numbers 2021769 and 2020250),
and by the
European Union (ERC,
VeriDeL, 101112713). Views and opinions expressed are however those of
the author(s) only and do not necessarily reflect those of the
European Union or the European Research Council Executive
Agency. Neither the European Union nor the granting authority can be
held responsible for them.
The work of Zhang was partially supported by the 
NSFC (grant number 62161146001).

% References will then be sorted \and formatted in the correct style.
%
\newpage
\bibliographystyle{splncs04}
\bibliography{marabou2}

\ifarxiv
\newpage
\appendix
\section{Architecture and Core Components}\label{app:sys}

\subsection{Supported Non-linear Constraints in \marabouTwo}\label{subapp:supportedconstraints}

\begin{table}
\small
\setlength\tabcolsep{4pt}
\centering
\begin{tabular}{lcccccc}
\toprule
\tabtitle{Type} & \tabtitle{Preproc.} & \tabtitle{NLR} & \tabtitle{SMT} & \tabtitle{(MI)LP} & \tabtitle{CEGAR} & \tabtitle{Proof} \\ 
\midrule
ReLU & \cmark & \cmark & \cmark & \cmark & \xmark & \cmark \\
Max  & \cmark & \cmark & \cmark & \cmark & \xmark & \cmark \\
DNF  & \cmark & \cmark & \cmark & \cmark & \xmark & \cmark \\
Sign  & \cmark & \cmark & \cmark & \cmark & \xmark & \cmark \\
Absolute value  & \cmark & \cmark & \cmark & \cmark & \xmark & \cmark \\
Leaky ReLU  & \cmark & \cmark & \cmark & \cmark & \xmark & \cmark \\
Round  & \cmark & \cmark & \cmark & \cmark & \xmark & \xmark \\
Clip  & \cmark & \cmark & \cmark & \cmark & \xmark & \xmark \\ 
Sigmoid  & \cmark & \cmark & \xmark & \cmark & \cmark & \xmark \\
Tanh  & \cmark & \cmark & \xmark & \cmark & \cmark & \xmark \\
Softmax  & \cmark & \cmark & \xmark & \cmark & \xmark & \xmark \\
Bi-linear  & \cmark & \cmark & \xmark & \cmark & \xmark & \xmark \\
\bottomrule
\end{tabular}
\end{table}

\ifnotarxiv
\section{Additional Use Cases of \marabou}\label{app:casestudies}

\fi

\section{\marabou in VNN-COMP 2023~\cite{brix2023fourth}}\label{app:vnn-comp}

\marabou participated in the Fourth International Verification of Neural Networks Competition (VNN-COMP'23) and won the second place overall and scored the highest among all CPU-based verifiers. Table~\ref{tab:stats1} shows the total number of instances solved by the participating solvers. Detailed information can be found in the VNN-COMP 2023 report~\cite{brix2023fourth}. The participating version is publicly available in the \marabou repo (commit \url{1a3ca6}). Overall, on the competition benchmarks, there is a non-trivial performance gap between \marabou and GPU-based solvers like $\alpha$-$\beta$-CROWN, especially for larger networks. Supporting GPUs in the Network-level Reasoner is on the development agenda for \marabou.

\begin{table}[ht]
\begin{center}
\caption{Total number of solved instances in VNN-COMP'23} \label{tab:stats1}
{\setlength{\tabcolsep}{4pt}
\begin{tabular}[h]{@{}llrc@{}}
\toprule
\tabtitle{\# ~} & \tabtitle{Tool} & \tabtitle{Count} & \tabtitle{GPU Support}\\
\midrule
1 & $\alpha$-$\beta$-CROWN & 721 & \cmark \\
2 & \marabou & 594 & \xmark \\
3 & NeuralSAT & 452 & \cmark \\
4 & PyRAT & 416 & \cmark\\
5 & nnenum & 319 & \xmark \\
6 & NNV & 155 & \xmark\\
7 & FastBATLLNN & 32 & \xmark \\
\bottomrule
\end{tabular}
}
\end{center}
\end{table}

\section{Runtime Evaluation}

\subsection{Descriptions of the Benchmarks}\label{subapp:benchmarks}

The \textbf{DeepCert} benchmarks come from a study~\cite{paterson2021deepcert} of the robustness of image classifiers against a set of analytically defined contextually relevant image perturbations. In particular, we focus on the haze perturbation on an image classifier (model 2A) for the GTSRB dataset. The \textbf{Alternating Loop} benchmarks come from a study of the existence of infinite loops in a robot navigation system~\cite{AmCoYeMaHaFaKa23}. The \textbf{NAP benchmarks} are less straightforward to procure: each benchmark considers an NAP mined on one of the ACAS-Xu networks by Gopinath et al.~\cite{gopinath2019property} and checks the \emph{robustness} of that NAP as defined by Geng et al.~\cite{geng2023towards}. We hope to also evaluate \marabou on the NAP robustness benchmarks described in the latter work, but they are not yet publicly available. Finally, the \textbf{VeriX} benchmarks involve performing a formal explanation extraction procedure (which repeatedly queries \marabou) on a MNIST classifier (the same one used in the original work).

\subsection{Solver Performance on Individual Benchmarks}\label{subapp:data}
\begin{center}
\setlength{\tabcolsep}{3pt}
\begin{xltabular}{\textwidth}{@{}llrrrrrr@{}}
\toprule
& & \multicolumn{3}{c}{\tabtitle{\marabouTwo}} & \multicolumn{3}{c}{\tabtitle{\marabou(1c1c66)}} \\
\cmidrule(lr){3-5} \cmidrule(lr){6-8}
Family & Benchmark  & Result & Time & Mem. & Result & Time & Mem. \\
\midrule
\endhead
Alt. Loops & REI.id114.ep67447& \sat & 0.6 & 6.6& \sat & 7.8 & 37.5\\
Alt. Loops & REI.id114.ep84297& \unsat & 55.7 & 10.6& \unsat & 1036.7 & 50.9\\
Alt. Loops & REI.id114.ep94159& \sat & 0.6 & 6.7& \sat & 22.3 & 42.6\\
Alt. Loops & REI.id114.ep94581& \sat & 0.6 & 6.7& \sat & 35.9 & 46.2\\
Alt. Loops & REI.id114.ep97993& \sat & 0.5 & 6.7& \sat & 139.3 & 48.4\\
Alt. Loops & REI.id128.ep68334& \unsat & 111.0 & 9.6& \unsat & 1749.2 & 56.0\\
Alt. Loops & REI.id128.ep85878& \unsat & 171.1 & 10.2& \unsat & 3478.2 & 60.4\\
Alt. Loops & REI.id128.ep87384& \sat & 0.7 & 6.3& \sat & 308.5 & 57.1\\
Alt. Loops & REI.id128.ep91042& \unsat & 94.9 & 9.2& \unsat & 1267.4 & 53.9\\
Alt. Loops & REI.id128.ep97983& \unsat & 127.2 & 10.0& \unsat & 2404.6 & 56.7\\
Alt. Loops & REI.id158.ep53113& \sat & 70.9 & 9.7& \sat & 16.9 & 34.8\\
Alt. Loops & REI.id158.ep53499& \sat & 14.6 & 8.6& \sat & 34.1 & 36.8\\
Alt. Loops & REI.id158.ep53554& \sat & 10.4 & 8.9& \sat & 17.7 & 39.0\\
Alt. Loops & REI.id158.ep59378& \unsat & 32.0 & 9.8& \unsat & 80.8 & 37.6\\
Alt. Loops & REI.id158.ep63427& \unsat & 23.7 & 9.4& \unsat & 68.9 & 38.7\\
Alt. Loops & REI.id165.ep82822& \sat & 0.7 & 6.8& \sat & 13.2 & 31.6\\
Alt. Loops & REI.id165.ep94134& \sat & 0.7 & 6.7& \sat & 14.4 & 33.9\\
Alt. Loops & REI.id165.ep94522& \sat & 0.5 & 6.5& \sat & 0.2 & 11.2\\
Alt. Loops & REI.id165.ep95976& \sat & 0.5 & 6.8& \sat & 30.7 & 34.6\\
Alt. Loops & REI.id165.ep96974& \sat & 24.0 & 8.3& \sat & 16.8 & 32.5\\
Alt. Loops & REI.id174.ep66258& \sat & 10.5 & 9.6& \sat & 0.1 & 5.8\\
Alt. Loops & REI.id174.ep81057& \sat & 0.6 & 6.5& \sat & 0.3 & 19.6\\
Alt. Loops & REI.id174.ep91107& \sat & 18.0 & 9.5& \sat & 49.7 & 43.4\\
Alt. Loops & REI.id174.ep97933& \sat & 0.7 & 6.7& \sat & 10.6 & 41.8\\
Alt. Loops & REI.id174.ep98455& \sat & 25.5 & 10.3& \sat & 0.2 & 13.0\\
Alt. Loops & REI.id176.ep92206& \unsat & 507.3 & 10.8& \unknown & 3601.0 & 66.0\\
Alt. Loops & REI.id176.ep95516& \unsat & 438.3 & 10.4& \unknown & 3601.0 & 61.5\\
Alt. Loops & REI.id176.ep96138& \unsat & 364.7 & 10.1& \unknown & 3601.0 & 63.1\\
Alt. Loops & REI.id176.ep98430& \unsat & 816.4 & 10.8& \unknown & 3601.0 & 64.6\\
Alt. Loops & REI.id176.ep99427& \unsat & 410.6 & 10.3& \unknown & 3601.0 & 64.2\\
Alt. Loops & REI.id180.ep58329& \sat & 0.5 & 6.2& \sat & 0.3 & 15.8\\
Alt. Loops & REI.id180.ep58461& \sat & 0.5 & 6.4& \sat & 0.2 & 14.3\\
Alt. Loops & REI.id180.ep61293& \sat & 0.6 & 6.4& \sat & 0.6 & 28.1\\
Alt. Loops & REI.id180.ep69795& \sat & 0.5 & 6.4& \sat & 0.2 & 9.6\\
Alt. Loops & REI.id180.ep77624& \sat & 0.4 & 6.6& \sat & 0.1 & 9.3\\
Alt. Loops & REI.id196.ep41633& \sat & 0.8 & 6.6& \sat & 0.2 & 12.0\\
Alt. Loops & REI.id196.ep43251& \sat & 0.3 & 6.4& \sat & 0.1 & 9.3\\
Alt. Loops & REI.id196.ep73462& \sat & 0.7 & 6.7& \sat & 0.2 & 12.8\\
Alt. Loops & REI.id196.ep86142& \sat & 0.3 & 6.4& \sat & 0.1 & 8.3\\
Alt. Loops & REI.id196.ep99092& \sat & 0.4 & 6.2& \sat & 24.0 & 34.0\\
Alt. Loops & REI.id201.ep39282& \sat & 21.5 & 9.6& \sat & 4.0 & 39.8\\
Alt. Loops & REI.id201.ep48186& \sat & 0.6 & 6.2& \sat & 51.4 & 43.1\\
Alt. Loops & REI.id201.ep48219& \sat & 0.6 & 6.4& \sat & 0.2 & 10.6\\
Alt. Loops & REI.id201.ep56910& \sat & 0.4 & 6.4& \sat & 28.7 & 34.9\\
Alt. Loops & REI.id201.ep88825& \sat & 0.2 & 6.0& \sat & 49.6 & 45.4\\
Alt. Loops & REI.id215.ep96277& \sat & 0.5 & 6.6& \sat & 1.7 & 41.1\\
Alt. Loops & REI.id215.ep97333& \sat & 0.3 & 6.4& \sat & 4.4 & 40.8\\
Alt. Loops & REI.id215.ep97355& \sat & 0.6 & 6.4& \sat & 0.6 & 32.9\\
Alt. Loops & REI.id215.ep97421& \sat & 0.5 & 6.9& \sat & 1763.0 & 60.1\\
Alt. Loops & REI.id215.ep98767& \sat & 0.5 & 6.5& \sat & 2.1 & 35.0\\
Alt. Loops & REI.id234.ep70556& \sat & 0.6 & 6.5& \sat & 0.1 & 10.8\\
Alt. Loops & REI.id234.ep71035& \sat & 0.3 & 6.2& \sat & 0.3 & 24.4\\
Alt. Loops & REI.id234.ep71965& \sat & 0.6 & 7.1& \sat & 0.2 & 11.8\\
Alt. Loops & REI.id234.ep83703& \sat & 0.7 & 6.7& \sat & 0.3 & 22.4\\
Alt. Loops & REI.id234.ep83780& \sat & 0.4 & 6.4& \sat & 1.0 & 33.8\\
Alt. Loops & REI.id239.ep71386& \unsat & 209.9 & 10.2& \unsat & 1691.7 & 55.5\\
Alt. Loops & REI.id239.ep73100& \unsat & 210.3 & 10.4& \unsat & 1110.3 & 60.0\\
Alt. Loops & REI.id239.ep73504& \sat & 6.7 & 9.1& \sat & 3.4 & 51.2\\
Alt. Loops & REI.id239.ep79112& \unsat & 193.2 & 9.9& \unsat & 1643.7 & 56.7\\
Alt. Loops & REI.id239.ep90118& \sat & 35.8 & 9.3& \sat & 0.4 & 18.0\\
Alt. Loops & REI.id240.ep85050& \unsat & 37.2 & 9.7& \unsat & 308.5 & 47.0\\
Alt. Loops & REI.id240.ep89317& \unsat & 26.5 & 9.4& \unsat & 152.7 & 42.7\\
Alt. Loops & REI.id240.ep91714& \unsat & 28.2 & 9.9& \unsat & 177.1 & 42.5\\
Alt. Loops & REI.id240.ep98115& \unsat & 40.2 & 10.2& \unsat & 282.4 & 47.6\\
Alt. Loops & REI.id240.ep98891& \unsat & 28.1 & 9.6& \unsat & 76.3 & 45.4\\
Alt. Loops & REI.id267.ep91426& \unsat & 0.7 & 6.2& \unsat & 0.7 & 14.5\\
Alt. Loops & REI.id267.ep95713& \unsat & 1.7 & 6.4& \unsat & 2.5 & 16.7\\
Alt. Loops & REI.id267.ep96300& \unsat & 1.7 & 6.9& \unsat & 1.0 & 14.8\\
Alt. Loops & REI.id267.ep96333& \unsat & 1.1 & 6.2& \unsat & 0.5 & 12.4\\
Alt. Loops & REI.id267.ep96426& \unsat & 1.3 & 6.0& \unsat & 1.5 & 15.5\\
Alt. Loops & REI.id286.ep89988& \unsat & 32.9 & 8.5& \unsat & 41.8 & 35.9\\
Alt. Loops & REI.id286.ep89999& \unsat & 32.4 & 8.5& \unsat & 41.7 & 35.8\\
Alt. Loops & REI.id286.ep90032& \unsat & 39.1 & 9.0& \unsat & 59.7 & 37.9\\
Alt. Loops & REI.id286.ep94925& \unsat & 35.1 & 9.2& \unsat & 56.8 & 38.1\\
Alt. Loops & REI.id286.ep95719& \unsat & 50.8 & 9.0& \unsat & 63.4 & 38.2\\
Alt. Loops & REI.id296.ep74854& \unsat & 140.7 & 9.7& \unsat & 697.8 & 51.9\\
Alt. Loops & REI.id296.ep74969& \unsat & 119.7 & 9.6& \unsat & 559.8 & 55.9\\
Alt. Loops & REI.id296.ep76414& \unsat & 127.1 & 10.1& \unsat & 500.1 & 51.7\\
Alt. Loops & REI.id296.ep77867& \unsat & 100.3 & 9.4& \unsat & 131.1 & 48.4\\
Alt. Loops & REI.id296.ep79125& \unsat & 133.1 & 9.9& \unsat & 274.0 & 46.4\\
Alt. Loops & REI.id298.ep75768& \unsat & 240.6 & 10.9& \unknown & 3601.0 & 66.3\\
Alt. Loops & REI.id298.ep77187& \sat & 0.8 & 6.8& \sat & 0.4 & 36.5\\
Alt. Loops & REI.id298.ep79143& \unsat & 167.6 & 10.9& \unknown & 3601.0 & 70.0\\
Alt. Loops & REI.id298.ep85589& \unsat & 175.5 & 10.8& \unknown & 3601.0 & 65.9\\
Alt. Loops & REI.id298.ep97154& \sat & 0.5 & 6.8& \sat & 0.1 & 8.0\\
Alt. Loops & REI.id299.ep69103& \unsat & 57.5 & 11.2& \unsat & 479.8 & 49.8\\
Alt. Loops & REI.id299.ep69400& \unsat & 49.6 & 10.7& \unsat & 361.7 & 47.0\\
Alt. Loops & REI.id299.ep81175& \unsat & 40.3 & 9.7& \unsat & 284.3 & 45.7\\
Alt. Loops & REI.id299.ep92705& \unsat & 111.3 & 10.7& \unsat & 739.8 & 50.7\\
Alt. Loops & REI.id299.ep99426& \unsat & 69.2 & 9.8& \unsat & 569.1 & 49.5\\
Alt. Loops & REI.id303.ep68520& \unsat & 243.1 & 11.6& \unknown & 3601.0 & 64.6\\
Alt. Loops & REI.id303.ep69708& \sat & 1.6 & 8.6& \sat & 987.6 & 61.3\\
Alt. Loops & REI.id303.ep94994& \sat & 0.6 & 6.6& \sat & 72.8 & 58.6\\
Alt. Loops & REI.id303.ep95038& \sat & 0.6 & 6.4& \sat & 4.4 & 52.3\\
Alt. Loops & REI.id303.ep95137& \sat & 0.4 & 6.4& \sat & 1399.1 & 72.3\\
Alt. Loops & REI.id308.ep94787& \unsat & 19.9 & 8.7& \unsat & 95.0 & 46.1\\
Alt. Loops & REI.id308.ep97397& \sat & 0.5 & 6.7& \sat & 0.2 & 14.4\\
Alt. Loops & REI.id308.ep98363& \sat & 0.7 & 6.6& \sat & 125.7 & 45.1\\
Alt. Loops & REI.id308.ep98830& \unsat & 52.5 & 10.1& \unsat & 207.0 & 46.9\\
Alt. Loops & REI.id308.ep99303& \sat & 0.8 & 6.9& \sat & 6.1 & 35.9\\
Alt. Loops & REI.id318.ep75036& \sat & 0.6 & 6.3& \sat & 14.4 & 64.8\\
Alt. Loops & REI.id318.ep85341& \sat & 96.4 & 10.7& \sat & 164.6 & 57.5\\
Alt. Loops & REI.id318.ep89536& \sat & 0.7 & 6.5& \sat & 27.3 & 55.8\\
Alt. Loops & REI.id318.ep94480& \sat & 0.5 & 6.5& \sat & 0.9 & 37.7\\
Alt. Loops & REI.id318.ep96308& \sat & 0.6 & 6.7& \sat & 0.4 & 21.7\\
Alt. Loops & REI.id319.ep62661& \sat & 0.4 & 6.5& \sat & 71.0 & 54.8\\
Alt. Loops & REI.id319.ep84434& \sat & 0.6 & 6.9& \sat & 2437.7 & 67.5\\
Alt. Loops & REI.id319.ep84478& \sat & 0.3 & 6.4& \sat & 0.3 & 14.3\\
Alt. Loops & REI.id319.ep86630& \sat & 0.5 & 6.2& \sat & 573.1 & 62.2\\
Alt. Loops & REI.id319.ep89649& \sat & 0.3 & 6.6& \sat & 0.3 & 12.9\\
Alt. Loops & REI.id321.ep55647& \unsat & 9.7 & 8.5& \unsat & 19.9 & 33.0\\
Alt. Loops & REI.id321.ep81083& \unsat & 18.2 & 8.5& \unsat & 404.1 & 53.6\\
Alt. Loops & REI.id321.ep86834& \unsat & 28.6 & 9.3& \unsat & 712.3 & 62.0\\
Alt. Loops & REI.id321.ep92994& \unsat & 25.0 & 9.5& \unsat & 389.5 & 54.3\\
Alt. Loops & REI.id321.ep93439& \unsat & 40.2 & 8.9& \unsat & 686.8 & 58.2\\
Alt. Loops & REI.id343.ep55239& \unsat & 62.6 & 9.2& \unsat & 149.7 & 47.1\\
Alt. Loops & REI.id343.ep63040& \unsat & 83.4 & 9.1& \unsat & 168.2 & 45.4\\
Alt. Loops & REI.id343.ep63051& \unsat & 83.2 & 9.1& \unsat & 166.6 & 45.5\\
Alt. Loops & REI.id343.ep73171& \unsat & 156.1 & 9.7& \unsat & 1022.7 & 54.2\\
Alt. Loops & REI.id343.ep96195& \unsat & 235.0 & 9.3& \unsat & 3113.7 & 57.7\\
Alt. Loops & REI.id35.ep61486& \sat & 0.4 & 6.4& \sat & 22.1 & 39.1\\
Alt. Loops & REI.id35.ep61788& \sat & 0.5 & 6.7& \sat & 0.2 & 10.9\\
Alt. Loops & REI.id35.ep81067& \sat & 0.3 & 6.3& \sat & 0.2 & 12.3\\
Alt. Loops & REI.id35.ep85842& \sat & 0.2 & 6.1& \sat & 0.2 & 10.4\\
Alt. Loops & REI.id35.ep94624& \sat & 0.4 & 6.4& \sat & 0.1 & 7.1\\
Alt. Loops & REI.id352.ep49376& \unsat & 7.6 & 8.2& \unsat & 58.8 & 39.3\\
Alt. Loops & REI.id352.ep51510& \unsat & 8.0 & 9.0& \unsat & 169.9 & 43.4\\
Alt. Loops & REI.id352.ep68171& \unsat & 12.1 & 9.2& \unsat & 87.1 & 39.8\\
Alt. Loops & REI.id352.ep69872& \unsat & 9.5 & 8.8& \unsat & 68.1 & 38.3\\
Alt. Loops & REI.id352.ep70312& \unsat & 8.1 & 8.2& \unsat & 24.0 & 37.8\\
Alt. Loops & REI.id379.ep71969& \sat & 1.4 & 7.9& \sat & 13.7 & 43.2\\
Alt. Loops & REI.id379.ep80359& \sat & 0.7 & 6.8& \sat & 0.2 & 11.7\\
Alt. Loops & REI.id379.ep87708& \unsat & 212.5 & 10.6& \unsat & 2425.8 & 57.3\\
Alt. Loops & REI.id379.ep96909& \sat & 32.1 & 10.7& \sat & 180.9 & 69.4\\
Alt. Loops & REI.id379.ep97133& \sat & 1.2 & 7.2& \sat & 0.7 & 36.4\\
Alt. Loops & REI.id381.ep85470& \sat & 1.9 & 9.2& \unknown & 3601.0 & 64.0\\
Alt. Loops & REI.id381.ep85558& \sat & 1.7 & 8.8& \sat & 1140.9 & 62.2\\
Alt. Loops & REI.id381.ep92541& \unsat & 250.5 & 11.2& \unknown & 3600.9 & 59.8\\
Alt. Loops & REI.id381.ep92994& \sat & 1.3 & 7.3& \unknown & 3601.0 & 64.2\\
Alt. Loops & REI.id381.ep93562& \unsat & 281.6 & 10.7& \unknown & 3601.0 & 64.3\\
Alt. Loops & REI.id393.ep88789& \sat & 25.5 & 11.0& \sat & 0.3 & 17.0\\
Alt. Loops & REI.id393.ep90550& \sat & 0.5 & 6.4& \sat & 64.9 & 56.5\\
Alt. Loops & REI.id393.ep97111& \sat & 31.6 & 10.0& \sat & 0.1 & 10.0\\
Alt. Loops & REI.id393.ep97199& \sat & 43.0 & 10.6& \sat & 0.2 & 15.2\\
Alt. Loops & REI.id393.ep97221& \sat & 37.3 & 11.0& \sat & 0.2 & 14.3\\
Alt. Loops & REI.id399.ep62611& \unsat & 42.9 & 9.2& \unsat & 490.9 & 56.3\\
Alt. Loops & REI.id399.ep69315& \unsat & 43.9 & 9.8& \unsat & 3342.7 & 68.6\\
Alt. Loops & REI.id399.ep85537& \unsat & 91.9 & 9.5& \unsat & 2220.3 & 60.8\\
Alt. Loops & REI.id399.ep92145& \unsat & 47.1 & 8.8& \unsat & 1475.9 & 64.6\\
Alt. Loops & REI.id399.ep96449& \sat & 1.5 & 7.1& \sat & 140.4 & 64.8\\
Alt. Loops & REI.id418.ep91524& \sat & 0.4 & 6.2& \sat & 5.3 & 53.1\\
Alt. Loops & REI.id418.ep92561& \sat & 0.6 & 6.6& \sat & 19.8 & 51.7\\
Alt. Loops & REI.id418.ep92671& \sat & 0.5 & 6.4& \sat & 30.0 & 49.7\\
Alt. Loops & REI.id418.ep99126& \sat & 0.9 & 6.4& \sat & 134.2 & 51.4\\
Alt. Loops & REI.id418.ep99947& \sat & 0.6 & 6.5& \sat & 13.4 & 51.0\\
Alt. Loops & REI.id425.ep69300& \sat & 0.6 & 6.7& \sat & 0.8 & 37.0\\
Alt. Loops & REI.id425.ep83340& \sat & 0.6 & 6.9& \sat & 0.3 & 14.0\\
Alt. Loops & REI.id425.ep93745& \unsat & 113.7 & 10.0& \unsat & 1299.1 & 51.4\\
Alt. Loops & REI.id425.ep96981& \sat & 0.5 & 6.6& \sat & 0.1 & 8.0\\
Alt. Loops & REI.id425.ep98177& \sat & 0.5 & 6.6& \sat & 0.1 & 11.8\\
Alt. Loops & REI.id444.ep77347& \unsat & 90.7 & 10.7& \unsat & 501.8 & 50.7\\
Alt. Loops & REI.id444.ep78487& \unsat & 120.5 & 10.2& \unsat & 1094.5 & 51.6\\
Alt. Loops & REI.id444.ep87148& \unsat & 78.3 & 10.0& \unsat & 842.5 & 51.1\\
Alt. Loops & REI.id444.ep92635& \unsat & 105.6 & 10.1& \unknown & 3601.0 & 48.8\\
Alt. Loops & REI.id444.ep96160& \unsat & 89.3 & 9.7& \unsat & 1262.5 & 51.4\\
Alt. Loops & REI.id457.ep46821& \sat & 11.0 & 9.3& \sat & 0.1 & 8.8\\
Alt. Loops & REI.id457.ep69348& \sat & 0.5 & 6.9& \sat & 0.2 & 10.6\\
Alt. Loops & REI.id457.ep78403& \sat & 0.3 & 6.4& \sat & 0.1 & 7.3\\
Alt. Loops & REI.id457.ep79411& \sat & 0.5 & 6.3& \sat & 0.2 & 12.7\\
Alt. Loops & REI.id457.ep81404& \sat & 0.6 & 6.5& \sat & 0.1 & 8.5\\
Alt. Loops & REI.id47.ep79665& \sat & 0.8 & 6.6& \sat & 0.3 & 19.2\\
Alt. Loops & REI.id47.ep92041& \sat & 0.3 & 6.3& \sat & 0.1 & 10.0\\
Alt. Loops & REI.id47.ep96122& \sat & 0.7 & 6.6& \sat & 0.4 & 21.0\\
Alt. Loops & REI.id47.ep99600& \sat & 0.6 & 6.4& \sat & 0.3 & 18.8\\
Alt. Loops & REI.id47.ep99688& \sat & 0.5 & 7.1& \sat & 0.3 & 18.2\\
Alt. Loops & REI.id491.ep54571& \sat & 0.3 & 6.6& \sat & 0.6 & 25.5\\
Alt. Loops & REI.id491.ep67504& \sat & 0.6 & 6.7& \sat & 1.3 & 33.2\\
Alt. Loops & REI.id491.ep67924& \sat & 0.4 & 6.6& \sat & 0.2 & 12.8\\
Alt. Loops & REI.id491.ep70134& \sat & 0.5 & 6.5& \sat & 0.4 & 23.2\\
Alt. Loops & REI.id491.ep96326& \sat & 0.4 & 6.1& \sat & 5.6 & 39.0\\
Alt. Loops & REI.id502.ep71027& \sat & 0.4 & 6.5& \sat & 56.7 & 42.9\\
Alt. Loops & REI.id502.ep71445& \sat & 0.4 & 6.5& \sat & 15.9 & 46.3\\
Alt. Loops & REI.id502.ep84337& \sat & 0.6 & 7.0& \sat & 2.1 & 30.2\\
Alt. Loops & REI.id502.ep86449& \sat & 0.8 & 6.9& \sat & 21.3 & 42.0\\
Alt. Loops & REI.id502.ep90122& \sat & 0.4 & 6.4& \sat & 266.6 & 45.8\\
Alt. Loops & REI.id512.ep70217& \sat & 0.6 & 6.6& \sat & 1.3 & 41.2\\
Alt. Loops & REI.id512.ep70283& \sat & 0.6 & 6.8& \sat & 1.0 & 33.3\\
Alt. Loops & REI.id512.ep75804& \sat & 0.5 & 6.4& \sat & 986.8 & 50.4\\
Alt. Loops & REI.id512.ep97330& \sat & 0.6 & 6.8& \sat & 0.1 & 11.2\\
Alt. Loops & REI.id512.ep97385& \sat & 0.4 & 6.3& \sat & 567.2 & 47.5\\
Alt. Loops & REI.id518.ep68431& \unsat & 97.2 & 9.9& \unsat & 523.8 & 46.8\\
Alt. Loops & REI.id518.ep70223& \unsat & 85.6 & 9.2& \unsat & 1358.8 & 49.7\\
Alt. Loops & REI.id518.ep95639& \unsat & 125.9 & 10.2& \unsat & 132.8 & 39.7\\
Alt. Loops & REI.id518.ep95900& \unsat & 123.8 & 9.9& \unsat & 602.4 & 45.1\\
Alt. Loops & REI.id518.ep96115& \unsat & 89.9 & 9.9& \unsat & 561.7 & 45.2\\
Alt. Loops & REI.id528.ep55711& \sat & 1.4 & 7.4& \sat & 1446.3 & 51.4\\
Alt. Loops & REI.id528.ep89177& \sat & 0.9 & 6.5& \sat & 53.8 & 53.1\\
Alt. Loops & REI.id528.ep89331& \sat & 0.8 & 6.9& \sat & 158.9 & 48.1\\
Alt. Loops & REI.id528.ep90341& \sat & 1.0 & 6.6& \sat & 36.3 & 39.8\\
Alt. Loops & REI.id528.ep95627& \sat & 0.5 & 6.3& \sat & 0.1 & 8.3\\
Alt. Loops & REI.id530.ep82130& \sat & 0.5 & 6.4& \sat & 0.2 & 13.2\\
Alt. Loops & REI.id530.ep82185& \sat & 0.5 & 6.7& \sat & 0.3 & 12.9\\
Alt. Loops & REI.id530.ep82207& \sat & 0.5 & 6.4& \sat & 0.2 & 13.1\\
Alt. Loops & REI.id530.ep91636& \sat & 0.5 & 6.8& \sat & 0.4 & 31.3\\
Alt. Loops & REI.id530.ep94463& \sat & 0.6 & 6.7& \sat & 0.2 & 12.8\\
Alt. Loops & REI.id535.ep76894& \unsat & 22.1 & 9.7& \unsat & 54.5 & 43.6\\
Alt. Loops & REI.id535.ep76916& \unsat & 58.5 & 9.3& \unsat & 51.4 & 39.6\\
Alt. Loops & REI.id535.ep87193& \sat & 0.8 & 6.8& \sat & 0.2 & 13.9\\
Alt. Loops & REI.id535.ep87226& \sat & 0.6 & 6.9& \sat & 0.5 & 22.8\\
Alt. Loops & REI.id535.ep91323& \unsat & 16.7 & 9.4& \unsat & 176.3 & 44.0\\
Alt. Loops & REI.id540.ep57595& \unsat & 1415.4 & 10.5& \unknown & 3601.0 & 65.2\\
Alt. Loops & REI.id540.ep57610& \unsat & 1984.2 & 10.8& \unknown & 3601.0 & 66.9\\
Alt. Loops & REI.id540.ep57775& \unsat & 1625.2 & 10.5& \unknown & 3601.0 & 68.2\\
Alt. Loops & REI.id540.ep85904& \sat & 0.8 & 6.8& \sat & 762.1 & 68.3\\
Alt. Loops & REI.id540.ep94690& \sat & 0.5 & 6.2& \sat & 0.1 & 12.2\\
Alt. Loops & REI.id549.ep85789& \sat & 0.5 & 6.6& \sat & 0.2 & 15.2\\
Alt. Loops & REI.id549.ep87887& \sat & 506.8 & 10.8& \sat & 1509.6 & 64.1\\
Alt. Loops & REI.id549.ep93366& \sat & 0.4 & 6.4& \sat & 28.6 & 51.1\\
Alt. Loops & REI.id549.ep99098& \sat & 1.0 & 6.8& \sat & 0.3 & 16.4\\
Alt. Loops & REI.id549.ep99331& \sat & 0.6 & 6.5& \sat & 0.1 & 8.2\\
Alt. Loops & REI.id67.ep82981& \unsat & 27.2 & 10.2& \unsat & 48.7 & 33.4\\
Alt. Loops & REI.id67.ep84585& \unsat & 53.2 & 10.6& \unsat & 98.8 & 39.6\\
Alt. Loops & REI.id67.ep87611& \unsat & 50.7 & 11.0& \unsat & 29.7 & 34.7\\
Alt. Loops & REI.id67.ep95115& \sat & 0.6 & 6.6& \sat & 111.2 & 41.5\\
Alt. Loops & REI.id67.ep99214& \sat & 0.6 & 6.8& \sat & 0.8 & 35.5\\
Alt. Loops & REI.id68.ep95777& \sat & 0.5 & 6.8& \sat & 0.4 & 20.0\\
Alt. Loops & REI.id68.ep95821& \sat & 0.6 & 6.5& \sat & 0.2 & 10.8\\
Alt. Loops & REI.id68.ep95854& \sat & 0.6 & 6.8& \sat & 0.3 & 16.7\\
Alt. Loops & REI.id68.ep95942& \sat & 0.5 & 6.4& \sat & 0.2 & 11.4\\
Alt. Loops & REI.id68.ep98804& \sat & 0.5 & 6.6& \sat & 0.3 & 23.0\\
Alt. Loops & REI.id73.ep74414& \sat & 328.7 & 11.0& \sat & 81.0 & 52.7\\
Alt. Loops & REI.id73.ep74469& \sat & 308.8 & 11.4& \sat & 23.7 & 50.4\\
Alt. Loops & REI.id73.ep82370& \unsat & 221.5 & 11.0& \unsat & 1166.3 & 58.1\\
Alt. Loops & REI.id73.ep84023& \unsat & 137.7 & 10.5& \unsat & 946.3 & 54.9\\
Alt. Loops & REI.id73.ep91504& \sat & 0.4 & 6.2& \sat & 153.7 & 53.7\\
Alt. Loops & REI.id76.ep90543& \unsat & 216.7 & 10.3& \unsat & 2095.2 & 62.3\\
Alt. Loops & REI.id76.ep91480& \unsat & 289.6 & 10.8& \unsat & 1264.9 & 62.4\\
Alt. Loops & REI.id76.ep92143& \unsat & 208.5 & 10.6& \unsat & 411.0 & 46.4\\
Alt. Loops & REI.id76.ep98692& \unsat & 310.5 & 11.0& \unsat & 1011.0 & 54.6\\
Alt. Loops & REI.id76.ep99198& \unsat & 315.0 & 10.9& \unsat & 1862.4 & 56.5\\
Alt. Loops & REI.id77.ep49509& \unsat & 25.0 & 8.8& \unsat & 69.4 & 37.8\\
Alt. Loops & REI.id77.ep75183& \unsat & 29.6 & 8.9& \unsat & 102.2 & 38.4\\
Alt. Loops & REI.id77.ep91446& \unsat & 52.7 & 9.6& \unsat & 165.0 & 41.0\\
Alt. Loops & REI.id77.ep99119& \unsat & 79.7 & 9.7& \unsat & 147.6 & 38.8\\
Alt. Loops & REI.id77.ep99229& \unsat & 73.5 & 8.9& \unsat & 144.3 & 40.5\\
Alt. Loops & REI.id81.ep89983& \unsat & 176.4 & 10.7& \unknown & 3601.0 & 59.4\\
Alt. Loops & REI.id81.ep91408& \unsat & 146.9 & 10.7& \unknown & 3601.0 & 60.9\\
Alt. Loops & REI.id81.ep91571& \unsat & 177.1 & 10.4& \unsat & 2485.5 & 56.8\\
Alt. Loops & REI.id81.ep97775& \unsat & 204.1 & 11.0& \unsat & 3578.0 & 61.4\\
Alt. Loops & REI.id81.ep97833& \unsat & 186.6 & 10.6& \unknown & 3601.0 & 57.0\\
Alt. Loops & REI.id90.ep84292& \unsat & 318.8 & 10.2& \unknown & 3601.0 & 60.4\\
Alt. Loops & REI.id90.ep94747& \unsat & 285.4 & 10.6& \unknown & 3601.0 & 59.0\\
Alt. Loops & REI.id90.ep97680& \unsat & 174.2 & 10.3& \unknown & 3601.0 & 54.8\\
Alt. Loops & REI.id90.ep97955& \unsat & 167.0 & 10.2& \unknown & 3601.0 & 59.9\\
Alt. Loops & REI.id90.ep99033& \unsat & 202.8 & 10.3& \unknown & 3601.0 & 58.0\\
Alt. Loops & REI.id91.ep54663& \unsat & 46.6 & 8.7& \unsat & 1790.1 & 53.0\\
Alt. Loops & REI.id91.ep58150& \unsat & 51.6 & 10.6& \unknown & 3601.0 & 55.4\\
Alt. Loops & REI.id91.ep58172& \unsat & 38.4 & 9.2& \unsat & 2933.8 & 56.0\\
Alt. Loops & REI.id91.ep75275& \unsat & 104.4 & 10.7& \unsat & 649.1 & 48.4\\
Alt. Loops & REI.id91.ep95105& \unsat & 125.5 & 10.0& \unsat & 1257.0 & 56.5\\
DeepCert & ind0.target6.eps0.2& \unsat & 8.7 & 399.7& \unsat & 268.9 & 603.7\\
DeepCert & ind0.target6.eps0.4& \unsat & 8.5 & 412.4& \unsat & 830.0 & 2053.0\\
DeepCert & ind0.target6.eps0.6& \unsat & 24.8 & 569.3& \unsat & 770.8 & 1252.2\\
DeepCert & ind14.target6.eps0.2& \unsat & 8.5 & 402.8& \unsat & 457.9 & 1081.1\\
DeepCert & ind14.target6.eps0.4& \sat & 26.9 & 561.2& \sat & 617.3 & 1421.2\\
DeepCert & ind14.target6.eps0.6& \sat & 24.3 & 569.9& \sat & 428.7 & 1086.1\\
DeepCert & ind15.target6.eps0.2& \unsat & 8.5 & 401.4& \unsat & 398.0 & 602.4\\
DeepCert & ind15.target6.eps0.4& \unsat & 39.4 & 560.2& \unsat & 476.5 & 951.0\\
DeepCert & ind15.target6.eps0.6& \sat & 24.4 & 568.7& \sat & 177.8 & 753.4\\
DeepCert & ind19.target6.eps0.2& \unsat & 8.5 & 402.8& \unsat & 442.5 & 938.9\\
DeepCert & ind19.target6.eps0.4& \unsat & 33.0 & 560.7& \unsat & 532.1 & 1419.9\\
DeepCert & ind19.target6.eps0.6& \sat & 25.5 & 569.5& \sat & 419.0 & 754.9\\
DeepCert & ind20.target6.eps0.2& \unsat & 8.6 & 403.3& \unsat & 352.1 & 603.2\\
DeepCert & ind20.target6.eps0.4& \unsat & 25.8 & 560.9& \unsat & 1568.0 & 2244.6\\
DeepCert & ind20.target6.eps0.6& \sat & 24.3 & 570.9& \sat & 436.5 & 1403.0\\
DeepCert & ind21.target6.eps0.2& \unsat & 8.7 & 402.4& \unsat & 366.8 & 601.6\\
DeepCert & ind21.target6.eps0.4& \unsat & 32.9 & 560.4& \unsat & 1054.4 & 1889.6\\
DeepCert & ind21.target6.eps0.6& \sat & 27.9 & 568.9& \sat & 874.6 & 2369.0\\
DeepCert & ind22.target6.eps0.2& \unsat & 8.6 & 403.0& \unsat & 261.3 & 602.2\\
DeepCert & ind22.target6.eps0.4& \unsat & 26.5 & 561.2& \unsat & 698.6 & 943.7\\
DeepCert & ind22.target6.eps0.6& \sat & 27.3 & 569.6& \sat & 376.1 & 3149.2\\
DeepCert & ind23.target6.eps0.2& \unsat & 8.8 & 401.9& \unsat & 615.3 & 1075.1\\
DeepCert & ind23.target6.eps0.4& \unsat & 33.7 & 560.4& \unsat & 727.4 & 1255.8\\
DeepCert & ind23.target6.eps0.6& \sat & 25.8 & 568.0& \sat & 162.7 & 910.2\\
DeepCert & ind24.target6.eps0.2& \unsat & 10.0 & 401.6& \unsat & 829.5 & 1259.0\\
DeepCert & ind24.target6.eps0.4& \unsat & 35.1 & 559.1& \unsat & 1477.3 & 1929.0\\
DeepCert & ind24.target6.eps0.6& \sat & 26.8 & 567.9& \sat & 272.1 & 933.5\\
DeepCert & ind25.target6.eps0.2& \unsat & 8.5 & 403.3& \unsat & 933.3 & 1572.4\\
DeepCert & ind25.target6.eps0.4& \sat & 24.6 & 560.6& \sat & 293.6 & 925.2\\
DeepCert & ind25.target6.eps0.6& \sat & 23.7 & 569.4& \sat & 357.4 & 1251.2\\
DeepCert & ind26.target6.eps0.2& \unsat & 8.6 & 404.0& \unsat & 515.5 & 781.7\\
DeepCert & ind26.target6.eps0.4& \unsat & 25.1 & 561.0& \unsat & 1136.0 & 1887.4\\
DeepCert & ind26.target6.eps0.6& \sat & 24.0 & 568.7& \sat & 1453.1 & 1568.0\\
DeepCert & ind27.target6.eps0.2& \unsat & 8.6 & 402.4& \unsat & 500.2 & 781.4\\
DeepCert & ind27.target6.eps0.4& \sat & 24.2 & 561.1& \sat & 628.1 & 1271.7\\
DeepCert & ind27.target6.eps0.6& \sat & 24.1 & 568.9& \sat & 905.0 & 2196.7\\
DeepCert & ind28.target6.eps0.2& \unsat & 8.5 & 402.4& \unsat & 714.5 & 945.7\\
DeepCert & ind28.target6.eps0.4& \unsat & 30.7 & 561.5& \unsat & 433.7 & 939.1\\
DeepCert & ind28.target6.eps0.6& \sat & 24.8 & 569.2& \sat & 437.0 & 1892.1\\
DeepCert & ind29.target6.eps0.2& \unsat & 8.4 & 402.1& \unsat & 597.4 & 782.6\\
DeepCert & ind29.target6.eps0.4& \sat & 23.6 & 561.2& \sat & 687.2 & 1112.0\\
DeepCert & ind29.target6.eps0.6& \sat & 25.2 & 569.3& \sat & 461.1 & 1248.9\\
DeepCert & ind31.target6.eps0.2& \unsat & 26.4 & 540.6& \unsat & 549.4 & 949.7\\
DeepCert & ind31.target6.eps0.4& \sat & 23.6 & 554.4& \sat & 309.5 & 1089.2\\
DeepCert & ind31.target6.eps0.6& \sat & 24.1 & 564.2& \sat & 366.0 & 1268.6\\
DeepCert & ind33.target6.eps0.2& \unsat & 20.9 & 539.6& \unsat & 804.9 & 1414.6\\
DeepCert & ind33.target6.eps0.4& \sat & 23.1 & 553.2& \sat & 211.3 & 1409.0\\
DeepCert & ind33.target6.eps0.6& \sat & 23.8 & 564.2& \sat & 357.8 & 757.1\\
DeepCert & ind39.target6.eps0.2& \unsat & 8.3 & 395.9& \unsat & 300.2 & 603.1\\
DeepCert & ind39.target6.eps0.4& \unsat & 34.5 & 553.4& \unsat & 1185.9 & 3033.1\\
DeepCert & ind39.target6.eps0.6& \unsat & 32.7 & 562.7& \unsat & 1742.9 & 2882.3\\
DeepCert & ind42.target6.eps0.2& \unsat & 8.4 & 396.2& \unsat & 603.8 & 940.5\\
DeepCert & ind42.target6.eps0.4& \unsat & 37.4 & 552.2& \unsat & 1143.6 & 3472.4\\
DeepCert & ind42.target6.eps0.6& \unsat & 56.1 & 561.8& \unsat & 1596.8 & 2047.7\\
DeepCert & ind43.target6.eps0.2& \unsat & 8.4 & 398.1& \unsat & 340.0 & 602.7\\
DeepCert & ind43.target6.eps0.4& \unsat & 29.6 & 555.1& \unsat & 1673.5 & 2244.1\\
DeepCert & ind43.target6.eps0.6& \unsat & 31.8 & 564.1& \unsat & 1622.2 & 3670.7\\
DeepCert & ind45.target6.eps0.2& \unsat & 8.4 & 397.3& \unsat & 428.6 & 781.9\\
DeepCert & ind45.target6.eps0.4& \unsat & 35.5 & 552.3& \unsat & 1208.6 & 2048.1\\
DeepCert & ind45.target6.eps0.6& \unsat & 25.6 & 562.3& \unsat & 2229.4 & 3658.6\\
DeepCert & ind46.target6.eps0.2& \unsat & 8.1 & 394.3& \unsat & 1287.0 & 1890.3\\
DeepCert & ind46.target6.eps0.4& \unsat & 24.6 & 548.8& \unsat & 1902.9 & 2328.9\\
DeepCert & ind46.target6.eps0.6& \unsat & 38.4 & 559.0& \unsat & 951.6 & 2042.0\\
DeepCert & ind47.target6.eps0.2& \unsat & 8.1 & 393.1& \unsat & 631.0 & 941.5\\
DeepCert & ind47.target6.eps0.4& \unsat & 24.4 & 547.9& \unsat & 736.4 & 1258.9\\
DeepCert & ind47.target6.eps0.6& \unsat & 30.7 & 557.4& \unsat & 1700.8 & 1771.6\\
DeepCert & ind51.target6.eps0.2& \unsat & 8.2 & 394.8& \unsat & 365.8 & 1102.2\\
DeepCert & ind51.target6.eps0.4& \unsat & 24.1 & 551.1& \unsat & 1807.4 & 2528.8\\
DeepCert & ind51.target6.eps0.6& \unsat & 40.6 & 559.9& \unsat & 2099.4 & 1973.8\\
DeepCert & ind53.target6.eps0.2& \unsat & 8.0 & 392.8& \unsat & 376.4 & 783.7\\
DeepCert & ind53.target6.eps0.4& \unsat & 24.5 & 547.4& \unsat & 1429.3 & 2523.9\\
DeepCert & ind53.target6.eps0.6& \unsat & 34.4 & 556.1& \unsat & 1480.0 & 1731.3\\
DeepCert & ind55.target6.eps0.2& \unsat & 7.9 & 389.9& \unsat & 235.3 & 600.9\\
DeepCert & ind55.target6.eps0.4& \unsat & 24.4 & 544.1& \unsat & 1388.7 & 1972.8\\
DeepCert & ind55.target6.eps0.6& \unsat & 30.8 & 553.0& \unsat & 875.4 & 2053.1\\
DeepCert & ind56.target6.eps0.2& \unsat & 8.0 & 391.0& \unsat & 579.3 & 1101.2\\
DeepCert & ind56.target6.eps0.4& \unsat & 23.5 & 544.3& \unsat & 809.9 & 1137.7\\
DeepCert & ind56.target6.eps0.6& \unsat & 38.8 & 553.4& \unsat & 1320.3 & 1728.9\\
DeepCert & ind6.target6.eps0.2& \unsat & 8.7 & 400.1& \unsat & 417.5 & 1098.1\\
DeepCert & ind6.target6.eps0.4& \unsat & 23.9 & 555.5& \unsat & 701.0 & 1576.8\\
DeepCert & ind6.target6.eps0.6& \sat & 26.1 & 564.1& \sat & 734.7 & 1255.1\\
DeepCert & ind60.target6.eps0.2& \unsat & 6.2 & 347.4& \unsat & 543.8 & 604.8\\
DeepCert & ind60.target6.eps0.4& \unsat & 6.0 & 358.8& \unsat & 302.9 & 599.8\\
DeepCert & ind60.target6.eps0.6& \unsat & 24.4 & 492.7& \unsat & 3366.6 & 2526.6\\
DeepCert & ind66.target6.eps0.2& \unsat & 5.1 & 321.8& \unsat & 465.4 & 601.4\\
DeepCert & ind66.target6.eps0.4& \sat & 14.5 & 443.9& \sat & 1079.6 & 1255.9\\
DeepCert & ind66.target6.eps0.6& \sat & 14.8 & 453.3& \sat & 1648.3 & 2406.2\\
DeepCert & ind67.target6.eps0.2& \unsat & 5.2 & 323.0& \unsat & 503.2 & 602.8\\
DeepCert & ind67.target6.eps0.4& \unsat & 18.6 & 447.2& \unsat & 994.9 & 1110.3\\
DeepCert & ind67.target6.eps0.6& \sat & 16.0 & 456.6& \sat & 2728.2 & 2375.9\\
DeepCert & ind68.target6.eps0.2& \unsat & 4.1 & 325.0& \unsat & 722.2 & 607.0\\
DeepCert & ind68.target6.eps0.4& \unsat & 4.4 & 303.8& \unsat & 679.7 & 949.3\\
DeepCert & ind68.target6.eps0.6& \unsat & 14.4 & 413.6& \unsat & 2760.5 & 1776.7\\
DeepCert & ind69.target6.eps0.2& \unsat & 4.9 & 308.4& \unsat & 583.5 & 604.9\\
DeepCert & ind69.target6.eps0.4& \unsat & 12.6 & 427.8& \unsat & 657.7 & 948.4\\
DeepCert & ind69.target6.eps0.6& \unsat & 15.7 & 435.7& \unsat & 3417.4 & 2102.0\\
DeepCert & ind7.target6.eps0.2& \unsat & 9.5 & 400.7& \unsat & 394.2 & 781.0\\
DeepCert & ind7.target6.eps0.4& \unsat & 26.4 & 558.2& \unsat & 1031.4 & 1734.9\\
DeepCert & ind7.target6.eps0.6& \sat & 26.6 & 567.1& \sat & 1410.3 & 2172.1\\
DeepCert & ind70.target6.eps0.2& \unsat & 5.1 & 310.3& \unsat & 721.9 & 604.4\\
DeepCert & ind70.target6.eps0.4& \unsat & 11.9 & 428.5& \unsat & 756.1 & 787.1\\
DeepCert & ind70.target6.eps0.6& \unsat & 17.1 & 437.9& \unsat & 2683.7 & 1617.1\\
DeepCert & ind71.target6.eps0.2& \unsat & 5.3 & 320.0& \unsat & 731.5 & 607.8\\
DeepCert & ind71.target6.eps0.4& \unsat & 5.0 & 330.5& \unsat & 1426.9 & 1421.8\\
DeepCert & ind71.target6.eps0.6& \unsat & 17.1 & 453.4& \unsat & 744.2 & 941.4\\
DeepCert & ind72.target6.eps0.2& \unsat & 5.3 & 330.3& \unsat & 536.4 & 604.2\\
DeepCert & ind72.target6.eps0.4& \unsat & 5.8 & 342.9& \unsat & 1011.7 & 1107.8\\
DeepCert & ind72.target6.eps0.6& \unsat & 16.3 & 469.0& \unsat & 934.7 & 944.2\\
DeepCert & ind73.target6.eps0.2& \unsat & 6.3 & 346.6& \unsat & 683.5 & 605.1\\
DeepCert & ind73.target6.eps0.4& \unsat & 6.1 & 359.2& \unsat & 826.2 & 942.2\\
DeepCert & ind73.target6.eps0.6& \unsat & 17.9 & 494.0& \unsat & 1616.3 & 1736.5\\
DeepCert & ind74.target6.eps0.2& \unsat & 6.4 & 351.5& \unsat & 722.7 & 606.0\\
DeepCert & ind74.target6.eps0.4& \unsat & 6.5 & 364.1& \unsat & 433.7 & 781.5\\
DeepCert & ind74.target6.eps0.6& \unsat & 19.6 & 500.8& \unsat & 1150.7 & 1734.0\\
DeepCert & ind75.target6.eps0.2& \unsat & 5.8 & 342.0& \unsat & 802.0 & 943.3\\
DeepCert & ind75.target6.eps0.4& \unsat & 5.9 & 353.1& \unsat & 423.9 & 601.3\\
DeepCert & ind75.target6.eps0.6& \unsat & 17.7 & 485.2& \unsat & 482.7 & 780.5\\
DeepCert & ind76.target6.eps0.2& \unsat & 5.8 & 339.9& \unsat & 705.4 & 604.7\\
DeepCert & ind76.target6.eps0.4& \unsat & 6.3 & 349.8& \unsat & 1254.2 & 941.2\\
DeepCert & ind76.target6.eps0.6& \unsat & 15.8 & 481.2& \unsat & 1929.3 & 1617.7\\
DeepCert & ind77.target6.eps0.2& \unsat & 5.8 & 342.9& \unsat & 638.5 & 604.6\\
DeepCert & ind77.target6.eps0.4& \unsat & 5.7 & 351.0& \unsat & 787.4 & 780.9\\
DeepCert & ind77.target6.eps0.6& \unsat & 5.9 & 359.5& \unsat & 1240.1 & 1452.1\\
DeepCert & ind78.target6.eps0.2& \unsat & 7.1 & 340.9& \unsat & 756.0 & 785.4\\
DeepCert & ind78.target6.eps0.4& \unsat & 6.7 & 351.8& \unsat & 556.1 & 603.6\\
DeepCert & ind78.target6.eps0.6& \unsat & 18.0 & 483.1& \unsat & 1791.8 & 2415.0\\
DeepCert & ind79.target6.eps0.2& \unsat & 7.1 & 353.6& \unsat & 566.5 & 603.6\\
DeepCert & ind79.target6.eps0.4& \unsat & 7.7 & 363.0& \unsat & 2526.8 & 1936.0\\
DeepCert & ind79.target6.eps0.6& \unsat & 20.8 & 498.7& \unsat & 2616.3 & 1935.7\\
DeepCert & ind8.target6.eps0.2& \unsat & 9.5 & 397.3& \unsat & 510.8 & 941.3\\
DeepCert & ind8.target6.eps0.4& \unsat & 34.3 & 552.4& \unsat & 1365.8 & 1495.2\\
DeepCert & ind8.target6.eps0.6& \sat & 24.7 & 561.8& \sat & 226.7 & 910.8\\
DeepCert & ind80.target6.eps0.2& \unsat & 7.9 & 367.0& \unsat & 527.5 & 604.0\\
DeepCert & ind80.target6.eps0.4& \unsat & 7.9 & 376.7& \unsat & 581.7 & 1257.7\\
DeepCert & ind80.target6.eps0.6& \unsat & 32.9 & 519.4& \unsat & 1696.3 & 1892.8\\
DeepCert & ind81.target6.eps0.2& \unsat & 7.5 & 360.0& \unsat & 355.7 & 602.4\\
DeepCert & ind81.target6.eps0.4& \unsat & 7.2 & 369.3& \unsat & 1711.7 & 2054.9\\
DeepCert & ind81.target6.eps0.6& \unsat & 26.3 & 508.4& \unsat & 3017.8 & 2612.1\\
DeepCert & ind82.target6.eps0.2& \unsat & 7.3 & 356.5& \unsat & 939.1 & 944.8\\
DeepCert & ind82.target6.eps0.4& \unsat & 8.0 & 365.3& \unsat & 848.6 & 1099.6\\
DeepCert & ind82.target6.eps0.6& \unsat & 20.4 & 502.4& \unsat & 1232.6 & 1573.7\\
DeepCert & ind83.target6.eps0.2& \unsat & 7.0 & 345.7& \unsat & 619.7 & 604.7\\
DeepCert & ind83.target6.eps0.4& \unsat & 7.0 & 354.2& \unsat & 794.1 & 781.3\\
DeepCert & ind83.target6.eps0.6& \unsat & 19.2 & 486.6& \unsat & 1028.2 & 1099.1\\
DeepCert & ind84.target6.eps0.2& \unsat & 6.3 & 337.1& \unsat & 541.6 & 603.8\\
DeepCert & ind84.target6.eps0.4& \unsat & 6.2 & 348.5& \unsat & 464.5 & 601.9\\
DeepCert & ind84.target6.eps0.6& \unsat & 17.6 & 478.1& \unsat & 1290.3 & 1742.3\\
DeepCert & ind9.target6.eps0.2& \unsat & 9.6 & 401.3& \unsat & 335.5 & 605.0\\
DeepCert & ind9.target6.eps0.4& \sat & 25.3 & 558.1& \sat & 380.7 & 757.5\\
DeepCert & ind9.target6.eps0.6& \sat & 26.2 & 567.5& \sat & 761.4 & 3307.1\\
NAP & cls0.id1& \sat & 2.7 & 43.2& \unknown & 3601.0 & 1002.2\\
NAP & cls0.id10& \unknown & 3601.0 & 57.1& \unknown & 3601.0 & 1170.2\\
NAP & cls0.id11& \sat & 1839.6 & 58.7& \unknown & 3601.0 & 1035.4\\
NAP & cls0.id12& \unknown & 3601.0 & 55.7& \unknown & 3601.0 & 1035.2\\
NAP & cls0.id13& \unknown & 3601.0 & 59.1& \unknown & 3601.0 & 1090.6\\
NAP & cls0.id14& \sat & 3.4 & 44.1& \unknown & 3601.0 & 952.8\\
NAP & cls0.id15& \unknown & 3601.0 & 57.5& \unknown & 3601.0 & 1139.4\\
NAP & cls0.id16& \unknown & 3601.0 & 59.2& \unknown & 3601.0 & 1074.2\\
NAP & cls0.id17& \sat & 179.7 & 56.8& \unknown & 3601.1 & 1272.1\\
NAP & cls0.id18& \unknown & 3601.0 & 57.1& \unknown & 3601.1 & 1106.1\\
NAP & cls0.id19& \unknown & 3601.0 & 56.6& \unknown & 3601.0 & 1132.6\\
NAP & cls0.id2& \unknown & 3601.0 & 56.4& \unknown & 3601.0 & 1050.2\\
NAP & cls0.id20& \unknown & 3601.0 & 58.8& \unknown & 3601.1 & 1192.9\\
NAP & cls0.id21& \sat & 2.3 & 42.8& \unknown & 3601.0 & 1197.1\\
NAP & cls0.id22& \unknown & 3601.0 & 57.1& \unknown & 3601.0 & 1024.7\\
NAP & cls0.id23& \unknown & 3601.0 & 56.5& \unknown & 3601.0 & 973.2\\
NAP & cls0.id24& \unknown & 3601.0 & 57.4& \unknown & 3601.1 & 1241.4\\
NAP & cls0.id25& \sat & 186.1 & 56.5& \unknown & 3601.0 & 1087.6\\
NAP & cls0.id26& \unknown & 3601.0 & 57.3& \unknown & 3601.0 & 984.7\\
NAP & cls0.id27& \unknown & 3601.0 & 56.0& \unknown & 3601.0 & 1095.6\\
NAP & cls0.id28& \unknown & 3601.0 & 57.0& \unknown & 3601.0 & 1039.9\\
NAP & cls0.id29& \unknown & 3601.0 & 56.7& \unknown & 3601.0 & 1034.3\\
NAP & cls0.id3& \sat & 1.7 & 43.2& \unknown & 3601.0 & 1219.9\\
NAP & cls0.id30& \sat & 209.4 & 58.0& \unknown & 3601.0 & 846.8\\
NAP & cls0.id31& \sat & 1583.9 & 57.1& \unknown & 3601.0 & 1058.1\\
NAP & cls0.id32& \unknown & 3601.0 & 57.2& \unknown & 3601.1 & 1208.1\\
NAP & cls0.id33& \sat & 57.8 & 55.6& \unknown & 3601.0 & 753.6\\
NAP & cls0.id34& \sat & 70.1 & 55.7& \unknown & 3601.0 & 970.2\\
NAP & cls0.id35& \unknown & 3601.0 & 58.3& \unknown & 3601.0 & 1094.9\\
NAP & cls0.id36& \unknown & 3601.0 & 57.4& \unknown & 3601.1 & 1128.7\\
NAP & cls0.id37& \unknown & 3601.0 & 55.8& \unknown & 3601.0 & 1008.3\\
NAP & cls0.id38& \unknown & 3601.0 & 56.7& \unknown & 3601.0 & 988.0\\
NAP & cls0.id39& \sat & 3.8 & 47.4& \unknown & 3601.1 & 1298.8\\
NAP & cls0.id4& \sat & 2.9 & 43.5& \unknown & 3601.0 & 1126.1\\
NAP & cls0.id40& \unknown & 3601.0 & 56.5& \unknown & 3601.0 & 1011.1\\
NAP & cls0.id41& \unknown & 3601.0 & 56.4& \unknown & 3601.0 & 1036.8\\
NAP & cls0.id42& \sat & 120.8 & 55.8& \unknown & 3601.1 & 1413.4\\
NAP & cls0.id43& \unknown & 3601.0 & 56.8& \unknown & 3601.1 & 1215.2\\
NAP & cls0.id44& \sat & 1415.6 & 56.9& \unknown & 3601.1 & 1372.7\\
NAP & cls0.id45& \unknown & 3601.0 & 57.1& \unknown & 3601.1 & 1435.9\\
NAP & cls0.id46& \unknown & 3601.0 & 57.4& \unknown & 3601.0 & 965.0\\
NAP & cls0.id47& \unknown & 3601.0 & 56.7& \unknown & 3601.1 & 1324.0\\
NAP & cls0.id48& \unknown & 3601.0 & 55.4& \unknown & 3601.0 & 1104.5\\
NAP & cls0.id49& \unknown & 3601.0 & 59.2& \unknown & 3601.0 & 928.7\\
NAP & cls0.id5& \sat & 2.0 & 43.8& \unknown & 3601.1 & 1152.9\\
NAP & cls0.id50& \sat & 403.0 & 55.9& \unknown & 3601.1 & 1138.8\\
NAP & cls0.id51& \unknown & 3601.0 & 57.2& \unknown & 3601.0 & 998.7\\
NAP & cls0.id52& \unknown & 3601.0 & 57.0& \unknown & 3601.1 & 1334.7\\
NAP & cls0.id53& \unknown & 3601.0 & 59.6& \unknown & 3601.0 & 1115.1\\
NAP & cls0.id54& \unknown & 3601.0 & 56.2& \unknown & 3601.1 & 1257.3\\
NAP & cls0.id55& \unknown & 3601.0 & 59.8& \unknown & 3601.0 & 846.1\\
NAP & cls0.id56& \unknown & 3601.0 & 57.1& \unknown & 3601.1 & 1247.0\\
NAP & cls0.id57& \unknown & 3601.0 & 57.2& \unknown & 3601.0 & 1078.0\\
NAP & cls0.id58& \unknown & 3601.0 & 58.9& \unknown & 3601.0 & 1090.7\\
NAP & cls0.id59& \unknown & 3601.0 & 57.7& \unknown & 3601.0 & 947.0\\
NAP & cls0.id6& \unknown & 3601.0 & 56.4& \unknown & 3601.0 & 1051.6\\
NAP & cls0.id60& \unknown & 3601.0 & 57.1& \unknown & 3601.0 & 970.5\\
NAP & cls0.id61& \unknown & 3601.0 & 55.5& \unknown & 3601.0 & 1106.2\\
NAP & cls0.id62& \unknown & 3601.0 & 58.0& \unknown & 3601.0 & 912.8\\
NAP & cls0.id63& \unknown & 3601.0 & 56.5& \unknown & 3601.0 & 1073.0\\
NAP & cls0.id64& \unknown & 3601.0 & 58.7& \unknown & 3601.0 & 937.1\\
NAP & cls0.id65& \unknown & 3601.0 & 57.9& \unknown & 3601.0 & 726.5\\
NAP & cls0.id66& \sat & 584.2 & 57.6& \unknown & 3601.0 & 1064.6\\
NAP & cls0.id67& \unknown & 3601.0 & 55.8& \unknown & 3601.1 & 1391.0\\
NAP & cls0.id68& \unknown & 3601.0 & 57.0& \unknown & 3601.0 & 956.8\\
NAP & cls0.id69& \unknown & 3601.0 & 56.2& \unknown & 3601.1 & 1249.9\\
NAP & cls0.id7& \sat & 1.6 & 43.8& \unknown & 3601.0 & 1106.7\\
NAP & cls0.id70& \unknown & 3601.0 & 56.8& \unknown & 3601.1 & 1110.0\\
NAP & cls0.id71& \unknown & 3601.0 & 56.2& \unknown & 3601.0 & 1027.2\\
NAP & cls0.id72& \unknown & 3601.0 & 56.9& \unknown & 3601.1 & 1166.1\\
NAP & cls0.id73& \sat & 1296.8 & 56.2& \unknown & 3601.0 & 1160.1\\
NAP & cls0.id74& \unknown & 3601.0 & 55.4& \unknown & 3601.1 & 962.4\\
NAP & cls0.id75& \unknown & 3601.0 & 55.9& \unknown & 3601.0 & 1137.9\\
NAP & cls0.id76& \unknown & 3601.0 & 55.4& \unknown & 3601.1 & 1317.1\\
NAP & cls0.id8& \unknown & 3601.0 & 57.4& \unknown & 3601.0 & 1043.1\\
NAP & cls0.id9& \sat & 2.3 & 43.1& \unknown & 3601.0 & 1193.6\\
NAP & cls1.id77& \unknown & 3601.0 & 59.9& \unknown & 3601.1 & 1149.2\\
NAP & cls1.id78& \unknown & 3601.0 & 59.4& \unknown & 3601.0 & 1102.7\\
NAP & cls1.id79& \sat & 5.2 & 50.7& \unknown & 3601.1 & 1157.8\\
NAP & cls1.id80& \unknown & 3601.0 & 62.0& \unknown & 3601.0 & 941.2\\
NAP & cls1.id81& \unknown & 3601.0 & 57.3& \unknown & 3601.0 & 981.8\\
NAP & cls1.id82& \unknown & 3601.0 & 62.1& \unknown & 3601.1 & 1297.8\\
NAP & cls1.id83& \unknown & 3601.0 & 58.7& \unknown & 3601.0 & 1002.1\\
NAP & cls1.id84& \sat & 1077.2 & 64.1& \unknown & 3601.0 & 1093.0\\
NAP & cls1.id85& \unknown & 3601.0 & 61.2& \unknown & 3601.0 & 1135.2\\
NAP & cls1.id86& \sat & 737.2 & 61.5& \unknown & 3601.0 & 837.2\\
NAP & cls1.id87& \unknown & 3601.0 & 58.3& \unknown & 3601.1 & 1222.5\\
NAP & cls1.id88& \unknown & 3601.0 & 57.2& \unknown & 3601.0 & 1068.2\\
NAP & cls2.id100& \unknown & 3601.0 & 57.0& \unknown & 3601.1 & 1246.2\\
NAP & cls2.id101& \unknown & 3601.0 & 59.1& \unknown & 3601.0 & 1035.8\\
NAP & cls2.id102& \unknown & 3601.0 & 57.6& \unknown & 3601.0 & 1107.1\\
NAP & cls2.id103& \unknown & 3601.0 & 57.2& \unknown & 3601.0 & 1088.9\\
NAP & cls2.id104& \unknown & 3601.0 & 56.4& \unknown & 3601.1 & 1166.8\\
NAP & cls2.id105& \unknown & 3601.0 & 55.5& \unknown & 3601.0 & 1220.3\\
NAP & cls2.id89& \unknown & 3601.0 & 58.3& \unknown & 3601.0 & 1030.3\\
NAP & cls2.id90& \unknown & 3601.0 & 59.1& \unknown & 3601.1 & 1142.3\\
NAP & cls2.id91& \unknown & 3601.0 & 55.6& \unknown & 3601.1 & 1096.6\\
NAP & cls2.id92& \unknown & 3601.0 & 58.4& \unknown & 3601.1 & 1205.4\\
NAP & cls2.id93& \unknown & 3601.0 & 60.4& \unknown & 3601.0 & 1057.7\\
NAP & cls2.id94& \unknown & 3601.0 & 58.4& \unknown & 3601.0 & 634.0\\
NAP & cls2.id95& \unknown & 3601.0 & 56.8& \unknown & 3601.0 & 881.4\\
NAP & cls2.id96& \unknown & 3601.0 & 56.5& \unknown & 3601.0 & 1049.3\\
NAP & cls2.id97& \unknown & 3601.0 & 56.5& \unknown & 3601.1 & 1166.9\\
NAP & cls2.id98& \unknown & 3601.0 & 62.1& \unknown & 3601.0 & 912.4\\
NAP & cls2.id99& \unknown & 3601.0 & 58.5& \unknown & 3601.1 & 1257.3\\
NAP & cls3.id106& \unknown & 3601.0 & 57.6& \unknown & 3601.0 & 899.9\\
NAP & cls3.id107& \sat & 371.1 & 58.8& \unknown & 3601.1 & 1267.5\\
NAP & cls3.id108& \unknown & 3601.0 & 63.3& \unknown & 3601.0 & 1093.3\\
NAP & cls3.id109& \unknown & 3601.0 & 57.1& \unknown & 3601.1 & 1428.3\\
NAP & cls3.id110& \unknown & 3601.0 & 62.0& \unknown & 3601.0 & 737.1\\
NAP & cls3.id111& \unknown & 3601.0 & 59.3& \unknown & 3601.0 & 1164.7\\
NAP & cls3.id112& \sat & 3146.8 & 63.5& \sat & 160.5 & 677.5\\
NAP & cls3.id113& \unknown & 3601.0 & 57.7& \unknown & 3601.0 & 1062.7\\
NAP & cls3.id114& \unknown & 3601.0 & 57.6& \unknown & 3601.1 & 1432.3\\
NAP & cls3.id115& \sat & 2.6 & 44.0& \unknown & 3601.1 & 1046.5\\
NAP & cls3.id116& \sat & 3.6 & 44.0& \unknown & 3601.0 & 1010.6\\
NAP & cls3.id117& \sat & 300.0 & 61.6& \unknown & 3601.0 & 931.4\\
NAP & cls3.id118& \sat & 13.2 & 53.9& \unknown & 3601.0 & 926.1\\
NAP & cls3.id119& \sat & 1019.9 & 61.1& \unknown & 3601.1 & 1147.0\\
NAP & cls3.id120& \unknown & 3601.0 & 59.4& \sat & 3323.4 & 1127.4\\
NAP & cls3.id121& \unknown & 3601.0 & 57.4& \unknown & 3601.0 & 802.9\\
NAP & cls3.id122& \unknown & 3601.0 & 61.2& \unknown & 3601.0 & 789.5\\
NAP & cls3.id123& \sat & 518.4 & 59.1& \unknown & 3601.0 & 1127.5\\
NAP & cls3.id124& \unknown & 3601.0 & 58.0& \unknown & 3601.1 & 1268.4\\
NAP & cls3.id125& \sat & 2.6 & 43.7& \unknown & 3601.1 & 1013.2\\
NAP & cls3.id126& \sat & 562.3 & 61.2& \unknown & 3601.1 & 1405.9\\
NAP & cls3.id127& \unknown & 3601.0 & 57.1& \unknown & 3601.1 & 1350.5\\
NAP & cls3.id128& \unknown & 3601.0 & 60.9& \unknown & 3601.1 & 1251.3\\
NAP & cls3.id129& \unknown & 3601.0 & 59.1& \unknown & 3601.0 & 1064.2\\
NAP & cls3.id130& \unknown & 3601.0 & 57.4& \unknown & 3601.0 & 886.6\\
NAP & cls3.id131& \unknown & 3601.0 & 61.4& \unknown & 3601.0 & 1146.7\\
NAP & cls3.id132& \sat & 2.8 & 44.3& \sat & 2882.8 & 955.8\\
NAP & cls3.id133& \unknown & 3601.0 & 62.7& \unknown & 3601.0 & 949.1\\
NAP & cls3.id134& \unknown & 3601.0 & 59.6& \unknown & 3601.1 & 1016.9\\
NAP & cls3.id135& \unknown & 3601.0 & 59.2& \unknown & 3601.0 & 1067.0\\
NAP & cls3.id136& \unknown & 3601.0 & 59.6& \unknown & 3601.0 & 928.2\\
NAP & cls3.id137& \unknown & 3601.0 & 58.6& \unknown & 3601.0 & 1060.5\\
NAP & cls3.id138& \unknown & 3601.0 & 56.6& \unknown & 3601.1 & 1055.6\\
NAP & cls3.id139& \unknown & 3601.0 & 57.3& \unknown & 3601.0 & 940.7\\
NAP & cls3.id140& \unknown & 3601.0 & 59.3& \sat & 2539.1 & 1029.8\\
NAP & cls3.id141& \unknown & 3601.0 & 60.7& \unknown & 3601.0 & 1010.3\\
NAP & cls3.id142& \unknown & 3601.0 & 59.7& \unknown & 3601.1 & 1276.4\\
NAP & cls3.id143& \unknown & 3601.0 & 57.2& \unknown & 3601.1 & 1107.5\\
NAP & cls3.id144& \unknown & 3601.0 & 57.3& \unknown & 3601.0 & 980.8\\
NAP & cls3.id145& \unknown & 3601.0 & 57.9& \sat & 81.3 & 905.3\\
NAP & cls3.id146& \unknown & 3601.0 & 61.9& \sat & 842.4 & 933.9\\
NAP & cls3.id147& \unknown & 3601.0 & 57.2& \unknown & 3601.1 & 1170.6\\
NAP & cls3.id148& \unknown & 3601.0 & 62.0& \unknown & 3601.0 & 1140.8\\
NAP & cls3.id149& \unknown & 3601.0 & 57.0& \unknown & 3601.0 & 978.0\\
NAP & cls3.id150& \unknown & 3601.0 & 57.2& \unknown & 3601.1 & 1595.4\\
NAP & cls3.id151& \unknown & 3601.0 & 57.7& \unknown & 3601.0 & 896.0\\
NAP & cls3.id152& \unknown & 3601.0 & 57.3& \unknown & 3601.0 & 1099.5\\
NAP & cls3.id153& \unknown & 3601.0 & 62.1& \unknown & 3601.0 & 995.1\\
NAP & cls3.id154& \unknown & 3601.0 & 57.3& \unknown & 3601.0 & 1167.0\\
NAP & cls3.id155& \unknown & 3601.0 & 57.3& \unknown & 3601.0 & 1098.2\\
NAP & cls3.id156& \unknown & 3601.0 & 55.1& \unknown & 3601.0 & 1009.3\\
NAP & cls3.id157& \unknown & 3601.0 & 57.0& \unknown & 3601.0 & 1106.1\\
NAP & cls3.id158& \unknown & 3601.0 & 57.1& \sat & 49.0 & 625.2\\
NAP & cls3.id159& \unknown & 3601.0 & 58.1& \unknown & 3601.0 & 1099.7\\
NAP & cls3.id160& \unknown & 3601.0 & 61.4& \unknown & 3601.0 & 963.7\\
NAP & cls3.id161& \unknown & 3601.0 & 59.3& \unknown & 3601.1 & 950.2\\
NAP & cls3.id162& \unknown & 3601.0 & 57.0& \unknown & 3601.0 & 1014.4\\
NAP & cls3.id163& \unknown & 3601.0 & 61.0& \unknown & 3601.0 & 1058.5\\
NAP & cls3.id164& \unknown & 3601.0 & 56.8& \unknown & 3601.0 & 1028.0\\
NAP & cls3.id165& \unknown & 3601.0 & 57.4& \unknown & 3601.1 & 1209.0\\
NAP & cls3.id166& \sat & 3.7 & 44.2& \unknown & 3601.1 & 1132.5\\
NAP & cls3.id167& \unknown & 3601.0 & 57.0& \unknown & 3601.0 & 962.5\\
NAP & cls3.id168& \sat & 1130.6 & 59.8& \unknown & 3601.1 & 1217.6\\
NAP & cls3.id169& \unknown & 3601.0 & 62.1& \unknown & 3601.1 & 1200.4\\
NAP & cls3.id170& \unknown & 3601.0 & 59.4& \unknown & 3601.0 & 820.0\\
NAP & cls3.id171& \unknown & 3601.0 & 61.2& \unknown & 3601.1 & 1468.8\\
NAP & cls3.id172& \unknown & 3601.0 & 57.0& \unknown & 3601.0 & 993.7\\
NAP & cls4.id173& \unknown & 3601.0 & 56.8& \unknown & 3601.0 & 907.6\\
NAP & cls4.id174& \unknown & 3601.0 & 58.7& \unknown & 3601.0 & 975.6\\
NAP & cls4.id175& \unknown & 3601.0 & 59.5& \unknown & 3601.1 & 1343.9\\
NAP & cls4.id176& \unknown & 3601.0 & 58.9& \unknown & 3601.0 & 1095.5\\
NAP & cls4.id177& \unknown & 3601.0 & 56.9& \unknown & 3601.1 & 1105.3\\
NAP & cls4.id178& \unknown & 3601.0 & 59.1& \unknown & 3601.0 & 982.6\\
NAP & cls4.id179& \unknown & 3601.0 & 56.8& \unknown & 3601.0 & 1010.6\\
NAP & cls4.id180& \sat & 1157.6 & 63.1& \unknown & 3601.1 & 1028.0\\
NAP & cls4.id181& \unknown & 3601.0 & 56.9& \unknown & 3601.1 & 1180.4\\
NAP & cls4.id182& \unknown & 3601.0 & 57.7& \sat & 171.3 & 530.2\\
NAP & cls4.id183& \unknown & 3601.0 & 55.0& \unknown & 3601.0 & 890.4\\
NAP & cls4.id184& \unknown & 3601.0 & 61.1& \unknown & 3601.0 & 968.7\\
NAP & cls4.id185& \unknown & 3601.0 & 60.8& \unknown & 3601.1 & 1083.6\\
NAP & cls4.id186& \unknown & 3601.0 & 57.3& \unknown & 3601.0 & 1089.6\\
NAP & cls4.id187& \unknown & 3601.0 & 56.9& \unknown & 3601.1 & 1143.1\\
NAP & cls4.id188& \unknown & 3601.0 & 58.2& \unknown & 3601.1 & 1324.8\\
NAP & cls4.id189& \unknown & 3601.0 & 57.5& \unknown & 3601.1 & 1257.3\\
NAP & cls4.id190& \unknown & 3601.0 & 56.5& \unknown & 3601.0 & 1255.5\\
NAP & cls4.id191& \unknown & 3601.0 & 57.8& \unknown & 3601.1 & 1081.5\\
NAP & cls4.id192& \unknown & 3601.0 & 56.1& \unknown & 3601.0 & 986.5\\
NAP & cls4.id193& \unknown & 3601.0 & 56.8& \unknown & 3601.1 & 1240.3\\
NAP & cls4.id194& \unknown & 3601.0 & 60.9& \unknown & 3601.0 & 824.6\\
NAP & cls4.id195& \sat & 2.3 & 44.5& \unknown & 3601.1 & 1105.0\\
NAP & cls4.id196& \unknown & 3601.0 & 59.2& \unknown & 3601.0 & 1081.9\\
NAP & cls4.id197& \unknown & 3601.0 & 56.5& \unknown & 3601.1 & 1350.3\\
NAP & cls4.id198& \sat & 2631.4 & 61.6& \unknown & 3601.1 & 1241.2\\
NAP & cls4.id199& \unknown & 3601.0 & 57.0& \unknown & 3601.0 & 793.0\\
NAP & cls4.id200& \unknown & 3601.0 & 57.6& \unknown & 3601.0 & 1015.7\\
NAP & cls4.id201& \unknown & 3601.0 & 59.1& \unknown & 3601.0 & 997.9\\
NAP & cls4.id202& \unknown & 3601.0 & 57.0& \unknown & 3601.0 & 812.3\\
NAP & cls4.id203& \unknown & 3601.0 & 60.1& \unknown & 3601.1 & 1084.9\\
NAP & cls4.id204& \unknown & 3601.0 & 56.4& \unknown & 3601.1 & 1514.7\\
NAP & cls4.id205& \unknown & 3601.0 & 60.5& \unknown & 3601.0 & 967.2\\
NAP & cls4.id206& \unknown & 3601.0 & 56.8& \unknown & 3601.0 & 949.8\\
NAP & cls4.id207& \unknown & 3601.0 & 56.7& \unknown & 3601.1 & 1294.3\\
NAP & cls4.id208& \unknown & 3601.0 & 56.7& \unknown & 3601.1 & 961.4\\
NAP & cls4.id209& \unknown & 3601.0 & 56.9& \unknown & 3601.1 & 1251.7\\
NAP & cls4.id210& \unknown & 3601.0 & 59.1& \unknown & 3601.0 & 774.3\\
NAP & cls4.id211& \unknown & 3601.0 & 57.1& \unknown & 3601.0 & 992.3\\
NAP & cls4.id212& \unknown & 3601.0 & 57.0& \unknown & 3601.1 & 1356.3\\
NAP & cls4.id213& \unknown & 3601.0 & 58.9& \unknown & 3601.0 & 1095.2\\
NAP & cls4.id214& \unknown & 3601.0 & 56.9& \unknown & 3601.1 & 1091.5\\
NAP & cls4.id215& \unknown & 3601.0 & 58.4& \unknown & 3601.1 & 1154.9\\
NAP & cls4.id216& \unknown & 3601.0 & 57.2& \unknown & 3601.1 & 1044.2\\
NAP & cls4.id217& \unknown & 3601.0 & 56.2& \unknown & 3601.0 & 887.3\\
NAP & cls4.id218& \unknown & 3601.0 & 58.2& \unknown & 3601.1 & 1367.5\\
NAP & cls4.id219& \unknown & 3601.0 & 58.7& \unknown & 3601.1 & 1085.1\\
NAP & cls4.id220& \unknown & 3601.0 & 56.3& \unknown & 3601.1 & 1167.6\\
NAP & cls4.id221& \unknown & 3601.0 & 57.1& \unknown & 3601.0 & 1051.8\\
NAP & cls4.id222& \unknown & 3601.0 & 58.9& \unknown & 3601.1 & 1407.6\\
NAP & cls4.id223& \sat & 267.4 & 61.3& \unknown & 3601.0 & 1072.7\\
NAP & cls4.id224& \unknown & 3601.0 & 57.7& \unknown & 3601.1 & 1041.2\\
NAP & cls4.id225& \unknown & 3601.0 & 56.0& \unknown & 3601.1 & 1136.6\\
NAP & cls4.id226& \unknown & 3601.0 & 54.0& \unknown & 3601.1 & 1185.4\\
NAP & cls4.id227& \unknown & 3601.0 & 58.8& \unknown & 3601.1 & 1159.9\\
NAP & cls4.id228& \unknown & 3601.0 & 56.3& \unknown & 3601.1 & 1199.0\\
NAP & cls4.id229& \unknown & 3601.0 & 56.7& \unknown & 3601.1 & 1493.2\\
NAP & cls4.id230& \unknown & 3601.0 & 56.4& \unknown & 3601.0 & 937.2\\
NAP & cls4.id231& \unknown & 3601.0 & 57.1& \unknown & 3601.0 & 822.6\\
NAP & cls4.id232& \unknown & 3601.0 & 57.3& \unknown & 3601.1 & 1199.7\\
NAP & cls4.id233& \unknown & 3601.0 & 56.9& \unknown & 3601.0 & 973.0\\
NAP & cls4.id234& \unknown & 3601.0 & 57.8& \sat & 15.6 & 265.8\\
NAP & cls4.id235& \unknown & 3601.0 & 56.0& \unknown & 3601.1 & 1155.8\\
VeriX & ind0& \ok & 571.7 & 778.7& \ok & 1483.1 & 1048.9\\
VeriX & ind1& \ok & 568.9 & 790.1& \ok & 1504.1 & 1045.5\\
VeriX & ind10& \ok & 670.7 & 528.1& \ok & 1665.6 & 501.9\\
VeriX & ind100& \ok & 1368.8 & 528.1& \unknown & 3601.0 & 502.0\\
VeriX & ind101& \ok & 3230.5 & 528.4& \unknown & 3601.0 & 503.0\\
VeriX & ind102& \ok & 630.0 & 529.6& \ok & 1683.7 & 502.6\\
VeriX & ind103& \ok & 989.4 & 527.7& \unknown & 3601.0 & 501.8\\
VeriX & ind11& \ok & 1456.7 & 529.3& \unknown & 3601.0 & 502.2\\
VeriX & ind12& \ok & 775.5 & 528.3& \ok & 2642.2 & 1048.7\\
VeriX & ind13& \ok & 1777.8 & 966.2& \unknown & 3601.0 & 502.0\\
VeriX & ind14& \ok & 674.8 & 1101.1& \ok & 1556.6 & 503.3\\
VeriX & ind15& \ok & 635.6 & 527.4& \ok & 1791.4 & 502.0\\
VeriX & ind16& \ok & 1210.9 & 527.6& \ok & 2818.4 & 501.8\\
VeriX & ind17& \ok & 543.7 & 527.9& \ok & 1449.7 & 502.3\\
VeriX & ind19& \ok & 2235.1 & 527.7& \unknown & 3601.0 & 502.1\\
VeriX & ind2& \ok & 1174.1 & 581.2& \ok & 3257.5 & 1045.6\\
VeriX & ind20& \ok & 667.8 & 528.5& \ok & 2091.4 & 1050.2\\
VeriX & ind21& \ok & 762.4 & 528.0& \ok & 2303.0 & 502.5\\
VeriX & ind22& \ok & 2646.4 & 528.5& \unknown & 3601.0 & 503.4\\
VeriX & ind23& \ok & 666.4 & 528.2& \ok & 1715.9 & 502.1\\
VeriX & ind24& \ok & 2386.4 & 528.4& \unknown & 3601.0 & 501.9\\
VeriX & ind25& \ok & 1662.5 & 528.8& \ok & 2197.4 & 502.5\\
VeriX & ind26& \ok & 759.5 & 528.1& \ok & 2392.8 & 502.4\\
VeriX & ind27& \ok & 982.9 & 528.0& \unknown & 3601.0 & 501.8\\
VeriX & ind28& \ok & 636.4 & 528.1& \ok & 1961.3 & 501.8\\
VeriX & ind29& \ok & 633.3 & 528.4& \ok & 1484.9 & 501.8\\
VeriX & ind3& \ok & 936.6 & 966.8& \ok & 2748.7 & 1045.5\\
VeriX & ind30& \ok & 667.1 & 528.3& \ok & 1703.9 & 502.7\\
VeriX & ind31& \ok & 1168.0 & 528.4& \ok & 1955.8 & 501.9\\
VeriX & ind32& \ok & 603.8 & 528.7& \ok & 1627.3 & 501.8\\
VeriX & ind34& \ok & 591.5 & 528.4& \ok & 1409.0 & 502.1\\
VeriX & ind35& \ok & 816.2 & 528.7& \ok & 1722.8 & 501.6\\
VeriX & ind36& \ok & 848.3 & 528.6& \ok & 1508.8 & 501.2\\
VeriX & ind37& \ok & 1212.0 & 528.5& \ok & 1942.0 & 502.3\\
VeriX & ind38& \ok & 429.9 & 528.2& \ok & 1100.8 & 502.4\\
VeriX & ind39& \ok & 1108.5 & 528.2& \ok & 2102.3 & 502.3\\
VeriX & ind4& \ok & 1615.9 & 1098.9& \unknown & 3601.0 & 502.1\\
VeriX & ind40& \unknown & 3601.0 & 529.6& \unknown & 3601.0 & 501.7\\
VeriX & ind41& \ok & 1381.1 & 528.1& \ok & 1641.0 & 502.0\\
VeriX & ind42& \ok & 737.1 & 529.1& \ok & 2096.1 & 502.0\\
VeriX & ind43& \ok & 1199.7 & 528.4& \ok & 3025.6 & 502.0\\
VeriX & ind44& \ok & 530.9 & 528.0& \ok & 1329.1 & 501.7\\
VeriX & ind45& \ok & 703.2 & 529.5& \ok & 1860.3 & 502.2\\
VeriX & ind46& \ok & 537.4 & 528.8& \ok & 1274.5 & 501.6\\
VeriX & ind47& \ok & 633.7 & 528.0& \ok & 1715.8 & 502.6\\
VeriX & ind48& \ok & 632.7 & 528.7& \ok & 1684.1 & 502.9\\
VeriX & ind49& \ok & 3228.2 & 528.2& \unknown & 3601.0 & 501.6\\
VeriX & ind5& \ok & 1135.8 & 1099.1& \ok & 3101.4 & 1047.2\\
VeriX & ind50& \ok & 1210.4 & 528.4& \ok & 2965.5 & 501.6\\
VeriX & ind51& \ok & 681.8 & 528.8& \ok & 1804.2 & 502.2\\
VeriX & ind52& \ok & 730.1 & 528.2& \ok & 1957.1 & 502.0\\
VeriX & ind53& \ok & 1615.6 & 529.1& \unknown & 3601.0 & 501.6\\
VeriX & ind54& \ok & 651.7 & 528.2& \ok & 1567.8 & 502.6\\
VeriX & ind55& \ok & 664.8 & 528.4& \ok & 1829.7 & 501.7\\
VeriX & ind56& \ok & 1014.9 & 528.1& \unknown & 3600.0 & 501.5\\
VeriX & ind57& \ok & 1403.4 & 528.8& \ok & 2715.4 & 501.6\\
VeriX & ind58& \ok & 718.6 & 528.9& \ok & 2200.0 & 502.4\\
VeriX & ind59& \ok & 3330.6 & 528.8& \unknown & 3601.0 & 501.6\\
VeriX & ind6& \ok & 655.4 & 528.6& \ok & 2558.8 & 1041.0\\
VeriX & ind60& \ok & 579.6 & 528.0& \ok & 1638.1 & 501.5\\
VeriX & ind61& \ok & 431.9 & 527.4& \ok & 1103.4 & 503.1\\
VeriX & ind62& \ok & 483.9 & 543.7& \ok & 1212.7 & 502.2\\
VeriX & ind64& \ok & 581.6 & 528.4& \ok & 1476.5 & 501.7\\
VeriX & ind65& \ok & 628.2 & 528.1& \ok & 2251.1 & 502.0\\
VeriX & ind66& \ok & 604.7 & 528.3& \ok & 2301.6 & 501.0\\
VeriX & ind67& \ok & 1415.0 & 528.9& \unknown & 3601.0 & 501.8\\
VeriX & ind68& \ok & 665.7 & 528.9& \ok & 1825.2 & 502.1\\
VeriX & ind69& \ok & 614.5 & 528.4& \ok & 1700.0 & 502.3\\
VeriX & ind7& \ok & 656.8 & 1092.9& \ok & 1854.0 & 509.2\\
VeriX & ind70& \ok & 1254.9 & 528.9& \ok & 1947.7 & 502.8\\
VeriX & ind71& \ok & 728.4 & 528.0& \ok & 1972.6 & 502.4\\
VeriX & ind72& \ok & 652.7 & 528.1& \ok & 1701.5 & 501.1\\
VeriX & ind73& \ok & 871.9 & 528.5& \ok & 2224.3 & 502.1\\
VeriX & ind74& \ok & 1096.8 & 528.2& \ok & 1902.1 & 501.9\\
VeriX & ind75& \ok & 826.6 & 527.8& \ok & 1841.3 & 501.3\\
VeriX & ind76& \ok & 647.8 & 528.0& \ok & 1898.4 & 501.2\\
VeriX & ind77& \ok & 3181.5 & 529.1& \unknown & 3601.0 & 502.6\\
VeriX & ind78& \ok & 631.6 & 545.2& \ok & 1777.7 & 501.9\\
VeriX & ind79& \ok & 595.8 & 528.0& \ok & 1527.9 & 502.3\\
VeriX & ind80& \ok & 618.7 & 544.5& \ok & 1633.8 & 502.4\\
VeriX & ind81& \ok & 762.3 & 528.3& \ok & 2279.7 & 502.1\\
VeriX & ind82& \ok & 633.9 & 528.1& \ok & 1741.3 & 502.1\\
VeriX & ind83& \ok & 693.7 & 527.9& \ok & 1831.7 & 501.8\\
VeriX & ind84& \ok & 1008.4 & 528.7& \ok & 3061.4 & 502.0\\
VeriX & ind85& \ok & 664.7 & 528.0& \ok & 1853.2 & 502.1\\
VeriX & ind86& \ok & 889.0 & 529.4& \ok & 1896.5 & 501.7\\
VeriX & ind87& \ok & 488.7 & 544.5& \ok & 1265.4 & 502.9\\
VeriX & ind88& \ok & 1373.3 & 527.7& \ok & 2453.0 & 501.4\\
VeriX & ind89& \ok & 1076.0 & 527.4& \ok & 2290.9 & 503.0\\
VeriX & ind9& \ok & 796.1 & 532.4& \ok & 2213.8 & 502.1\\
VeriX & ind90& \ok & 649.4 & 529.3& \ok & 1842.2 & 502.9\\
VeriX & ind91& \ok & 754.0 & 528.4& \ok & 2204.0 & 501.3\\
VeriX & ind92& \ok & 589.6 & 528.5& \ok & 1653.2 & 501.5\\
VeriX & ind93& \ok & 617.3 & 527.3& \ok & 1608.4 & 502.3\\
VeriX & ind94& \ok & 692.8 & 528.5& \ok & 1637.0 & 502.8\\
VeriX & ind95& \ok & 614.4 & 529.1& \ok & 1863.5 & 503.2\\
VeriX & ind96& \ok & 1197.3 & 528.0& \ok & 1689.1 & 502.4\\
VeriX & ind97& \ok & 373.8 & 545.5& \ok & 979.6 & 501.7\\
VeriX & ind98& \ok & 600.4 & 528.7& \ok & 1500.1 & 501.9\\
VeriX & ind99& \ok & 1290.3 & 527.9& \ok & 3130.3 & 501.8 \\
\bottomrule
\end{xltabular}

\end{center}

\fi

\end{document}

%%% Local Variables:
%%% mode: latex
%%% TeX-master: t
%%% End: